\documentclass{article}

\usepackage[utf8]{inputenc}
\usepackage[T1]{fontenc}

\usepackage{microtype}
\usepackage{graphicx}
\usepackage{booktabs}

\usepackage{hyperref}

\usepackage{amsmath}
\usepackage{amssymb}
\usepackage{amsfonts}

\usepackage[accepted]{icml2025}

\makeatletter
\renewcommand{\ICML@appearing}{Preprint.}
\renewcommand{\Notice@String}{Preprint.}
\makeatother

\hypersetup{hypertexnames=false}

\icmltitlerunning{The Anatomy of the CTC Oracle Gap}

\begin{document}

\twocolumn[
\icmltitle{The Anatomy of the CTC Oracle Gap:\\
           Acoustic Exhaustion and Linguistic Recovery}

\begin{icmlauthorlist}
\icmlauthor{Ivan Novosad}{hse}
\end{icmlauthorlist}

\icmlaffiliation{hse}{Faculty of Computer Science, HSE University, Moscow, Russia}

\icmlcorrespondingauthor{Ivan Novosad}{inovosad@hse.ru}

\icmlkeywords{Connectionist Temporal Classification, Minimum Bayes Risk decoding,
pseudo-log-likelihood, Rao-Blackwellization, sequence-level ASR training,
N-best rescoring, word error rate, cross-condition generalization}

\vskip 0.3in
]

\printAffiliationsAndNotice{}

\hypersetup{
  pdftitle={The Anatomy of the CTC Oracle Gap: Acoustic Exhaustion and Linguistic Recovery},
  pdfauthor={Ivan Novosad},
  pdfsubject={},
  pdfkeywords={CTC, MBR decoding, pseudo-log-likelihood, ASR, N-best rescoring},
}

\begin{abstract}

This paper investigates the limits of CTC-internal scoring for N-best hypothesis selection and identifies the information bottleneck that separates acoustic confidence from linguistic plausibility.

Eleven CTC-internal and acoustic-feature scoring strategies — including MBR with CTC posteriors, MC-dropout, contrastive decoding, shallow fusion, and a trained encoder value head — produce no statistically significant WER improvement over greedy decoding on LibriSpeech dev-other at $G$=16 (all $p > 0.05$). The exhaustion is systematic: CTC's Spearman $\rho$ between hypothesis score and per-utterance WER degrades from $-$0.574 at $G$=4 to $-$0.270 at $G$=128, a 53\% loss in ranking quality driven by blank-path proliferation in larger candidate sets. This exhaustion result establishes that the discriminative capacity of CTC-internal representations is saturated — no reweighting or recombination of acoustic signals can close the oracle gap.

As confirmation that the bottleneck is linguistic rather than acoustic, we show that introducing external linguistic information via MBR decoding breaks through it. MBR-CER decoding with a RoBERTa pseudo-log-likelihood posterior ($\tau$=10, $G$=128) achieves 5.42\% WER on held-out LibriSpeech test-other (greedy 5.96\%, $\Delta$=$-$0.535 pp, $p$$<$0.0001, 9.0\% relative). RoBERTa PLL $\rho$ degrades only 21\% over the same $G$=4$\to$128 range, maintaining discriminating power exactly where CTC loses it. Applied without retuning to two architectures (Zipformer-S/M), three domains (LibriSpeech, TED-LIUM 3, VoxPopuli), and four noise levels (MUSAN), the recipe produces statistically significant improvements in 11 of 13 conditions, with VoxPopuli and MUSAN at 0\,dB as the two predicted failure conditions where candidate quality degrades.

On the training-time side, we establish that standard MWER training via the CTC forward-backward algorithm implements Rao-Blackwellized REINFORCE at the output projection (variance ${\sim}3\times$ below Viterbi, verified on 250 utterances). Despite this variance reduction, sequence-level fine-tuning fails at near-converged CTC checkpoints: all four MWER configurations on CR-CTC collapse (+6.18 to +8.90 pp WER) due to a training oracle gap of 0.007 pp that provides no usable reward signal.


\end{abstract}

\begin{figure*}[t]
  \centering
  \includegraphics[width=\textwidth]{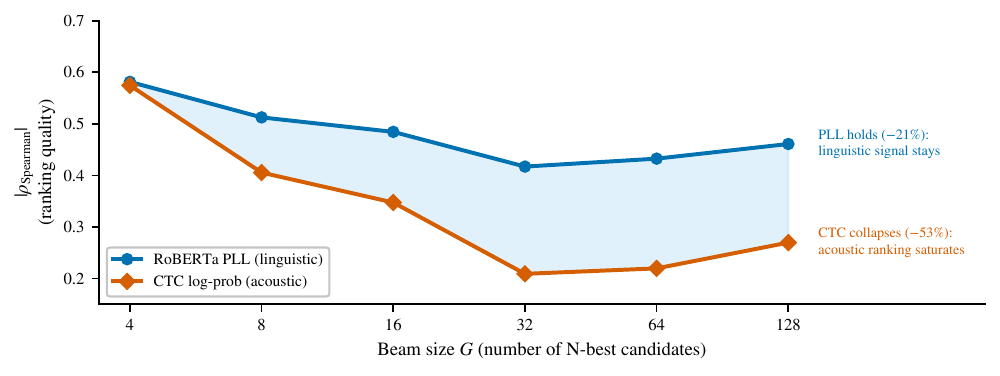}
  \caption{\textbf{Where the CTC oracle gap lives.} Ranking quality (absolute
  Spearman $\rho$ between scorer and per-utterance WER) versus beam size $G$ on
  LibriSpeech dev-other. CTC log-probability degrades sharply as the candidate
  set grows (blank-path proliferation), while RoBERTa pseudo-log-likelihood (PLL)
  stays informative. This divergence is the mechanism behind the paper's central
  finding: the near-converged CTC decoding bottleneck is \emph{linguistic, not
  acoustic}. MBR decoding with a PLL posterior exploits it for a 9.0\% relative
  WER reduction on held-out test-other.}
  \label{fig:teaser}
\end{figure*}

\section{Introduction}
\label{sec:intro}

\subsection{Problem Domain}

Automatic speech recognition based on Connectionist Temporal Classification \cite{graves2006ctc} trains without frame-level alignment supervision and achieves competitive word error rates while decoding quickly. A CTC model outputs a distribution over vocabulary tokens at each encoder frame; the probability of a token sequence $y$ given acoustic input $x$ is computed by marginalizing over all valid frame-to-token alignments via the forward-backward algorithm. This marginalization is exact and runs in $O(T \cdot |y|)$ time. Conditional independence across output frames enables parallel decoding, and the model family competes with attention-based encoder-decoder systems while remaining simpler to deploy in latency-sensitive settings.

Greedy decoding selects the argmax token at each encoder frame, then collapses consecutive duplicates and blanks. It is fast and exact in the per-frame MAP sense. It does not maximize the true sequence probability $P_\text{CTC}(y|x)$ after alignment marginalization: the per-frame argmax tokens need not compose into the most probable token sequence. The distance between greedy decoding and the best decoding the model can theoretically support is not negligible.

On LibriSpeech dev-other \cite{panayotov2015librispeech}, a 22M-parameter Zipformer-S model \cite{yao2024zipformer} trained with the CR-CTC objective \cite{huang2024crctc} achieves 6.02\% WER under greedy decoding. The \textbf{oracle WER} is the WER of the best candidate per utterance in the N-best list; the \textbf{oracle gap} is greedy WER minus oracle WER — the maximum WER reduction achievable by perfect candidate selection. Drawing $G=16$ candidates via k2 lattice sampling \cite{li2022k2} yields an oracle WER of 4.44\% — an oracle gap of 1.58~pp. At $G=128$, the oracle reaches 3.53\%, an oracle gap of 2.49~pp. These are deterministic properties of the model and dataset, computed once with fixed parameters. The oracle gap is stable and reproducible. At $G=128$, it is 2.49~pp — larger than the absolute WER improvement this paper reports — which establishes both the scale of the opportunity and the difficulty of fully closing it.

Better transcripts exist in the N-best list. Two strategies exist for finding them. The first is training: use a sequence-level objective that rewards WER reduction to modify the model parameters, exploiting the oracle gap as the available reward signal. The second is decoding: hold the model fixed and use information not available to the acoustic model to select a better candidate from the N-best list. CTC's separation of acoustic scoring from transcript selection makes both strategies feasible on the same frozen model.

Both strategies are applied here to the same Zipformer-S CR-CTC checkpoint. Training fails; decoding succeeds. Section~\ref{sec:training} identifies why; Sections~\ref{sec:decoding} and~\ref{sec:cross_condition} document how much ground decoding recovers and where it runs out.

\subsection{Informal Problem Statement}

This paper asks a precise question: given a frozen CTC acoustic model, can WER be reduced by (a) sequence-level training on the model parameters, or (b) decision-theoretic decoding over the model's N-best output? The underlying analytical question is diagnostic: where does the information bottleneck lie — within the acoustic model's own representations, or at the boundary between acoustic and linguistic domains?

On the training side, the RL framing provides analytical tools. A CTC model defines a distribution $P_\text{CTC}(y|x)$ over token sequences, which in the RL vocabulary is a stochastic policy $\pi_\theta(y|x)$ parameterized by model weights $\theta$. The REINFORCE algorithm \cite{williams1992} gives the gradient of expected reward as $\mathbb{E}_{y \sim \pi_\theta}\bigl[R(y)\,\nabla_\theta \log \pi_\theta(y)\bigr]$. Setting $R(y) = -\text{WER}(y, y^*)$ and approximating the expectation with a finite N-best list recovers the Minimum Word Error Rate (MWER) loss of \citet{prabhavalkar2018mwer}; mean-centered WER advantages over the list serve as the reward baseline. The functional-form identification is exact; the Rao-Blackwell variance property is established empirically in §3. Section~\ref{sec:theory} extends it: the CTC forward-backward algorithm that computes $\nabla_\theta \log P_\text{CTC}(y|x)$ implements a Rao-Blackwellized REINFORCE estimator, automatically providing variance reduction that any MWER practitioner already obtains without explicitly invoking the theorem.

At decode time, the RL framing does not apply. With model parameters fixed, there is no policy to optimize. The problem is action selection under uncertainty — Bayesian decision theory. Minimum Bayes Risk (MBR) decoding \cite{kumar2004mbr} provides the formal solution: select the hypothesis minimizing expected loss under a posterior distribution over the N-best candidates. When the posterior incorporates external linguistic evidence, the MBR solution differs from the acoustic argmax. This is decision theory, not RL. Section~\ref{sec:theory} draws this boundary explicitly.

Two questions follow: why does sequence-level training fail on near-converged CTC checkpoints, and — more diagnostically — what does the failure of CTC-internal scoring at decode time reveal about where the model's discriminative capacity ends?

\subsection{Relevance and Novelty}

LM rescoring for CTC is well-studied. Shallow fusion \cite{gulcehre2015fusion} combines acoustic and LM log-probabilities at each beam expansion step; applied to attention-based encoder-decoder ASR, it produces consistent WER reductions across languages and model sizes \cite{hori2017advances}. WFST-based decoding \cite{miao2015eesen} integrates n-gram models into the CTC search graph as a compiled grammar. Within the n-gram model class, this yields exact joint search. N-best rescoring \cite{mikolov2010recurrent,shin2019effective} runs an LM offline over a fixed candidate list and reranks by combined score. All three approaches assign a single value to each candidate and select the argmax; the decoding objective remains MAP.

MBR decoding changes the objective. \citet{kumar2004mbr} introduce MBR for machine translation and show gains over MAP decoding on multiple benchmarks; the improvement is largest where probability mass spreads across near-equivalent hypotheses. \citet{eikema2020map} formalize the argument: MAP decoding is systematically misaligned with the output goal when the model distribution is diffuse. \citet{freitag2022high} scale MBR with learned neural utility functions, reporting substantial gains on NMT benchmarks. For ASR, \citet{goel2000mbr} apply MBR to acoustic lattices in hybrid HMM systems, but the posterior in that work derives entirely from the acoustic model and carries no external linguistic signal. The closest precedent for this work is \citet{finkelstein2024mbr} in NMT, which replaces the LM score with a neural loss function $L$; we change the posterior $Q$ itself and operate in the CTC ASR domain.

\citet{salazar2020pll} introduce pseudo-log-likelihood (PLL) for masked language model scoring: each token position is individually masked, and the sum of per-position log-probabilities under the masked contexts provides a sentence-level score. As \citet{salazar2020pll} show, PLL correlates with human grammaticality judgments and outperforms autoregressive scores on acceptability benchmarks. Prior ASR work uses PLL as a drop-in score in an argmax pipeline, selecting the highest-PLL candidate. This work uses PLL to construct the MBR posterior: PLL defines the probability distribution over candidates from which MBR computes expected loss.

The combination of masked-LM posteriors with an MBR objective over CTC N-best candidates does not appear in the prior literature. Three findings emerge from this investigation.

\textit{(1) Exhaustion of CTC-internal scoring.} Eleven scoring strategies derivable from CTC posteriors, encoder features, or their combinations produce no statistically significant WER improvement over greedy. The Spearman $\rho$ divergence between CTC and PLL ranking quality as beam size grows provides the mechanistic explanation: CTC's discriminative capacity degrades under blank-path proliferation while PLL remains stable. This establishes that the bottleneck is linguistic, not acoustic.

\textit{(2) External information breaks the bottleneck.} MBR decoding with RoBERTa \cite{liu2019roberta} PLL as the posterior and CER as the loss function achieves a 9.0\% relative WER reduction on held-out test-other. The same recipe transfers without modification to two architectures (Zipformer-S and Zipformer-M), three domains (LibriSpeech, TED-LIUM~3 \cite{hernandez2018tedlium3}, VoxPopuli \cite{wang2021voxpopuli}), and four noise levels (MUSAN \cite{snyder2015musan}), reaching significance in 11 of 13 conditions.

\textit{(3) Analysis of MWER failure.} A $2\times2$ outcome matrix covers two model types (CR-CTC and standard CTC) and two training objectives (MWER and supervised distillation). All four cells fail; the analysis identifies two distinct failure mechanisms and establishes the boundary conditions under which sequence-level fine-tuning at near-converged CTC checkpoints could succeed.

\subsection{Results Summary}

\textbf{Training-time (Section~\ref{sec:training}).} All four MWER training configurations on the CR-CTC checkpoint degrade WER monotonically: degradation ranges from $+6.18$ to $+8.90$~pp on dev-other. The cause is a near-zero training oracle gap of 0.007~pp — the model already transcribes training data near-perfectly, so the REINFORCE gradient tracks noise rather than a reward signal. MWER on a standard CTC checkpoint does not produce catastrophic collapse but still fails: a mild upward drift of $+3.4\%$ over 3000 steps, despite a training oracle gap 106$\times$ larger than CR-CTC's. Best-of-N supervised distillation (RAFT) on the same checkpoint collapses at lr$=10^{-6}$ (WER rises from 4.86\% to 65.5\% by step 800) and is frozen at lr$=10^{-7}$ and lr$=10^{-8}$; the collapse boundary lies within one order of magnitude. The behavior is consistent with a sharp basin in the loss surface around the pretrained optimum. The $2\times2$ outcome matrix in Section~\ref{sec:training} summarizes: two model types, two training objectives, four cells, no positive results, two distinct failure mechanisms.

\textbf{Decode-time (Sections~\ref{sec:decoding}–\ref{sec:cross_condition}).} Eleven CTC-internal and acoustic-feature scoring strategies produce no statistically significant WER improvement at $G=16$ (all $p > 0.05$). The information bottleneck lies between CTC and the linguistic domain. MBR-CER decoding with a RoBERTa PLL posterior at temperature $\tau=10$ achieves 5.42\% WER on the held-out LibriSpeech test-other split at $G=128$ (greedy 5.96\%, $\Delta=-0.535$~pp, $p < 0.0001$, 95\% CI~$[-0.629,\,-0.441]$~pp). This is a 9.0\% relative reduction from the greedy baseline, confirmed on held-out data with no hyperparameter adjustment. The effect strengthens on test-other relative to dev-other, a pattern inconsistent with development-set overfitting. The result confirms the information-bottleneck hypothesis: when external linguistic information is introduced, the CTC-internal ceiling no longer holds.

The absolute number alone does not capture cross-condition transferability. The same recipe — temperature $\tau=10$, CER utility, $G=128$ — applies without modification to two Zipformer architectures, three domains, and four additive noise levels. Statistically significant improvements appear in 11 of 13 conditions tested. The two failures are VoxPopuli (coverage collapse: 91.5\% of utterances already greedy-optimal) and MUSAN at 0~dB (candidate quality degrades under extreme noise). Both failures are predicted by the mechanistic analysis in Section~\ref{sec:decoding} before those experiments ran. The recipe requires no per-condition adaptation; this transferability across the 11 conditions where candidate quality permits is the claim this paper defends. The master results table in Section~\ref{sec:cross_condition} presents the full evidence.

\subsection{Paper Structure}

Related work (§2) surveys CTC decoding, LM rescoring, MBR in MT and ASR, and sequence-level training; the specific combination motivating this work does not appear in any of these threads. The theoretical framework (§3) derives the CTC alignment posterior $\gamma_t$, identifies MWER training as Rao-Blackwellized REINFORCE (variance ${\sim}3\times$ below Viterbi, verified on 250 utterances), and verifies two variance-bounding propositions on up to 2{,}864 dev-other utterances before making the transition from training-time analysis to decode-time decision theory.

The training experiments (§4) are organized around a $2\times2$ outcome matrix; §5 documents the exhaustion of CTC-internal scoring, introduces external LM posteriors, and explains the Spearman $\rho$ divergence behind the MBR-vs-interpolation scaling asymmetry; §6 validates the recipe across 13 conditions and reports four diagnostics. §7 synthesizes the findings.

\section{Related Work}

\subsection{CTC Decoding and N-best Generation}

CTC \cite{graves2006ctc} trains an acoustic model to assign probability to token sequences without frame-level alignment supervision. The model outputs a distribution $p_t(k | x)$ over the vocabulary at each encoder frame $t$; the probability of a token sequence $y$ is obtained by marginalizing over all valid frame-to-token alignments via the forward-backward algorithm, which runs in $O(T \cdot |y|)$ time. Each frame-level prediction depends on the full acoustic context through the encoder but not on adjacent output decisions. This conditional independence at the output projection enables parallel decoding and makes CTC well-suited to latency-sensitive settings.

Greedy decoding selects the argmax label at each frame and collapses consecutive duplicates and blanks. It is fast and exact in the MAP sense over frame-level decisions, but it does not maximize the true CTC sequence probability $P_\text{CTC}(y|x)$: the per-frame argmax labels do not necessarily compose into the highest-probability token sequence after alignment marginalization. Prefix beam search \cite{graves2014towards} corrects this by maintaining a set of partial hypotheses and accumulating CTC probabilities over all valid alignments for each prefix. Prefix beam search produces a scored N-best list ordered by accumulated CTC probability and remains the standard CTC decoding algorithm.

A single beam search does not guarantee diverse candidates. When the model is confident, the top-$G$ hypotheses from a fixed beam may differ only in a few tokens, and the oracle WER of the N-best list barely undercuts the greedy WER. The k2 framework \cite{li2022k2} addresses this through lattice-based sampling: a first-pass lattice is built over a wide beam, and $G$ hypotheses are drawn from it using a temperature parameter (nbest\_scale) that controls arc-weight sharpness. At nbest\_scale=1.0, the oracle WER at $G=128$ on LibriSpeech dev-other reaches 3.53\%, against a greedy baseline of 6.02\% — a 2.49~pp oracle gap. Reducing nbest\_scale to 0.5 destroys approximately 90\% of this gap by collapsing the lattice distribution prematurely; all experiments in this work use nbest\_scale=1.0.

The primary acoustic model throughout this paper is a Zipformer-S \cite{yao2024zipformer} trained with the CR-CTC objective \cite{huang2024crctc}. CR-CTC augments the CTC loss with a consistency regularization term: the KL divergence between the output posteriors of two stochastically augmented views of the same utterance is added as a penalty. The effect on the alignment posterior $\gamma_t$ is measurable. On 100 dev-other utterances, the CR-CTC model produces a near-equal three-way split — dead frames (blank-dominated, $\gamma_t(\text{blank}) > 0.99$) at 34.2\%, active frames at 34.7\%, and ambiguous frames at 31.1\%. Unregularized CTC models concentrate 70–90\% of frames in the dead category \cite{graves2006ctc}; CR-CTC roughly halves that fraction by spreading alignment mass more evenly across active frames.

CTC posterior scores alone cannot close this gap: CTC scoring tracks acoustic confidence rather than linguistic plausibility, and the ranking signal degrades as the beam grows. Section~\ref{sec:exhaustion} verifies this across eleven CTC-internal scoring strategies, all of which fail to produce statistically significant WER improvements. An external linguistic signal is required.

\subsection{Language Model Rescoring for ASR}

Two strategies dominate LM integration in end-to-end ASR: shallow fusion and N-best rescoring. Shallow fusion \cite{gulcehre2015fusion} combines the acoustic model log-probability with an autoregressive LM log-probability at each decoding step:
\begin{equation*}
\log \tilde{P}(y_t | y_{<t}, x) \;=\; \log P_\text{AM}(y_t | x) \;+\; \lambda\, \log P_\text{LM}(y_t | y_{<t}).
\end{equation*}
The interpolation weight $\lambda$ is tuned on a held-out set. \citet{hori2017advances} apply this approach to attention-based encoder-decoder ASR and demonstrate consistent WER reductions across languages and model sizes. The strategy is straightforward to implement, but its limitation is that the LM must be queried at every beam expansion step, coupling inference latency to LM size.

N-best rescoring decouples acoustic decoding from LM scoring. The acoustic decoder runs first and produces a fixed candidate list; the LM scores each candidate offline, and the combined score determines the final selection. \citet{mikolov2010recurrent} show that recurrent neural network LMs substantially outperform n-gram models in this setting. Their work establishes the second-pass neural rescoring paradigm that modern systems still follow. \citet{shin2019effective} bring this to end-to-end ASR: a pre-trained transformer LM scoring the CTC N-best list achieves WER reductions comparable to shallow fusion without modifying the acoustic decoder architecture.

For CTC systems, WFST-based decoding \cite{miao2015eesen} offers a third option. The CTC decoding graph is composed with a compiled n-gram language model in a weighted finite-state transducer; the result enables joint search over acoustic and language model scores. Within the n-gram model class, this approach is exact. Its structural limitation is that the grammar must be compiled at a fixed vocabulary and n-gram order, so on-the-fly updates and neural LM integration are impractical.

All three approaches share a common form for the LM score: an autoregressive or n-gram conditional probability $P(y_t | y_{<t})$. Bidirectional masked LMs such as RoBERTa \cite{liu2019roberta} produce non-autoregressive posteriors that do not fit this interface and cannot be used in shallow fusion or WFST integration. \citet{salazar2020pll} introduce pseudo-log-likelihood (PLL) as a sentence-level score for masked LMs: each token position is individually masked, and the sum of per-position log-probabilities under the masked contexts defines the PLL for the complete sentence. PLL correlates with human grammaticality judgments and outperforms autoregressive LM scores on acceptability benchmarks \cite{salazar2020pll}. In prior ASR work, however, PLL serves only as a drop-in replacement for conventional LM scores in a rescoring-and-argmax pipeline.

This work changes the decoding objective rather than the scoring function. The PLL score defines a probability distribution over the N-best candidates, and MBR decoding then selects the hypothesis minimizing expected CER under that distribution — not the hypothesis with the highest score. When the CTC posterior concentrates mass around near-duplicate candidates, the argmax and MBR solutions diverge; the MBR solution tends to be better in these cases, as the Spearman-$\rho$ divergence analysis in §5.3 quantifies. To our knowledge, the combination of masked-LM posteriors with an MBR objective applied to CTC N-best lists has not been explored before this work.

\subsection{Minimum Bayes Risk Decoding}

MBR decoding selects the output that minimizes expected loss under a posterior over the hypothesis space:
\begin{equation*}
\hat{y}_\text{MBR} \;=\; \operatorname{argmin}_{y \in \mathcal{H}} \sum_{y' \in \mathcal{H}} Q(y') \, L(y, y'),
\end{equation*}
where $\mathcal{H}$ is the candidate set, $Q(y')$ is a probability distribution over hypotheses, and $L$ is a task-specific loss function. MAP decoding selects the hypothesis assigned highest model probability, without accounting for the cost structure $L$ — that is, without considering how wrong each candidate would be if each alternative turned out to be correct. \citet{kumar2004mbr} introduce the MBR formulation for machine translation and show that sentence-level BLEU MBR outperforms MAP decoding on multiple benchmarks; the gain is largest on outputs where several near-equivalent hypotheses share high model probability.

In machine translation, MBR has become a standard technique. \citet{eikema2020map} argue theoretically that MAP decoding is ill-suited to neural sequence generation: the probability mass of a well-trained model spreads across many paraphrase variants of the same meaning, so any single MAP hypothesis is unrepresentative of the distribution. Sampling-based MBR, which draws a large hypothesis set from the model and applies the MBR objective over the sample, produces more fluent and accurate translations. \citet{freitag2022high} scale this approach using learned neural metrics (BLEURT, COMET) as the utility function $L$ and report substantial improvements over beam search on WMT benchmarks. \citet{finkelstein2024mbr} extend the framework to training: NMT models fine-tuned to minimize expected loss produce better translations than cross-entropy-trained models; the result shows that MBR and the training objective can be unified.

MBR for ASR has precedent, though it predates the neural era. \citet{goel2000mbr} apply a minimum expected WER criterion to acoustic lattice rescoring in hybrid HMM-based systems: the posterior $Q(y')$ is estimated from the acoustic lattice, and the hypothesis minimizing expected WER under this posterior is selected in place of the MAP hypothesis. These results show consistent improvements over MAP decoding on the NIST Switchboard benchmark. In that work, however, $Q(y')$ is derived entirely from the acoustic model. No external linguistic signal enters the posterior; the lattice already encodes all available information.

When $Q(y')$ is the CTC acoustic distribution, MBR over the N-best list can only reweight candidates within what the acoustic model already knows. The Spearman rank correlation between CTC hypothesis scores and per-utterance WER degrades from $-0.347$ at $G=16$ to $-0.270$ at $G=128$ as near-duplicate candidates dilute the acoustic ranking signal — a 22\% drop. PLL degrades much more gracefully over the same range ($-0.484$ to $-0.461$, a 5\% drop), and the two signals diverge most sharply on utterances where the correct transcription differs substantially from the CTC argmax. Augmenting the posterior with PLL changes which candidates MBR selects in precisely these recoverable cases. The closest precedent is \citet{finkelstein2024mbr} in NMT, which uses neural metrics as the utility function $L$ while retaining an autoregressive model posterior; this paper changes the posterior $Q$ itself and operates in the ASR domain.

\subsection{RL and Sequence-Level Training for ASR}

The connection between policy gradient RL and sequence-level ASR training is well established. \citet{williams1992} introduces REINFORCE: the gradient of expected reward with respect to policy parameters equals the expected product of the reward and the gradient of the log-policy probability. Applied to sequence generation, the action is the output token sequence, the reward is a sequence-level metric such as WER, and the policy is the acoustic model distribution. \citet{prabhavalkar2018mwer} derive the MWER training loss directly from this framing:
\begin{equation*}
\mathcal{L}_\text{MWER} \;=\; \sum_{i=1}^{G} \hat{A}_i \;\log P_\text{CTC}(y_i|x),
\end{equation*}
where $\hat{A}_i = \text{WER}(y_i, y^*) - \frac{1}{G}\sum_{j} \text{WER}(y_j, y^*)$ is the mean-centered WER advantage over the N-best list. \citet{shannon2017} develop the same functional form and show that sequence-level fine-tuning after cross-entropy pretraining consistently reduces WER. Both works treat the REINFORCE identification as motivation rather than as an object of analysis.

REINFORCE gradient estimates have high variance, which motivates several reduction strategies. \citet{rennie2017self} propose self-critical sequence training (SCST): the reward baseline is the performance of the model's own greedy output at test time. This adaptive baseline generalizes the mean-reward baseline of \citet{williams1992} by conditioning on the actual test-time decoding decision. The Rao-Blackwell theorem \cite{casella1996rao} provides a more structural guarantee: replacing a random variable with its conditional expectation given a sufficient statistic produces an estimator with weakly lower variance. \citet{ranganath2014bbvi} apply Rao-Blackwellization to black-box variational inference by conditioning score-function gradient estimators on subsets of the latent variables. This reduces per-sample variance without changing the expected gradient.

Recent work on large language models revisits these ideas with importance-ratio clipping. GRPO \cite{shao2024grpo} extends REINFORCE with a group-relative advantage baseline and a PPO-style clipping term, improving training stability for LLM fine-tuning on reasoning benchmarks. The structure is the same as MWER: a finite action space, a non-differentiable reward, and a policy that is already near-converged before RL training begins. Section~\ref{sec:training} applies clipping of this type to the CR-CTC setting; it reduces the magnitude of degradation but does not reverse its direction.

The theoretical contribution of this work connects these ideas through the CTC backward pass. Standard MWER training computes the gradient of $\log P_\text{CTC}(y_i|x)$ via the CTC forward-backward algorithm. This returns the alignment posterior $\gamma_t(k|y_i)$ as the per-frame gradient weight. We show in Section~\ref{sec:theory} that this marginalization over alignments is exactly a Rao-Blackwellization of a single-alignment REINFORCE estimator: $\gamma_t(k|y_i)$ is the conditional expectation of the Viterbi or sampled per-frame indicator given the token sequence $y_i$, and the Rao-Blackwell theorem guarantees its variance is weakly lower. Empirically, the CTC-marginal gradient has 2.96$\times$ lower variance than the Viterbi estimator and 3.87$\times$ lower variance than a sampled single-path estimator. Any practitioner running standard MWER training via the CTC backward pass already obtains this variance reduction; the contribution of this paper is identifying and verifying the mechanism. The RL framing is limited to training-time optimization: at decode time, the model parameters are fixed, and the problem becomes one of action selection under uncertainty — Bayesian decision theory, developed in Section~\ref{sec:theory}.

\section{Theoretical Framework}
\label{sec:theory}

The framework splits at the training–decoding boundary: Sections 3.1–3.3 cover the training side — the CTC alignment posterior $\gamma_t$, its Rao-Blackwell connection to MWER, and two variance-bounding propositions verified on LibriSpeech dev-other — while Section 3.4 marks the crossing point from training-time analysis to decode-time decision theory.

\subsection{CTC and the Alignment Posterior $\gamma_t$}

The Connectionist Temporal Classification loss \cite{graves2006ctc} trains a model to assign probability to a token sequence $y$ given acoustic features $x \in \mathbb{R}^{T \times d}$, without requiring frame-level alignment labels. Let $p_t(k|x)$ denote the model's output probability for token $k$ at frame $t$, for $k$ ranging over the vocabulary $\mathcal{V}$ of size $V$ plus a designated blank symbol. A CTC alignment is a sequence $\pi \in \mathcal{V}^T$ that collapses, via the many-to-one map $\mathcal{B}$ (remove consecutive duplicates, then remove blanks), to the token sequence $y$. The CTC probability of $y$ marginalizes over all valid alignments:

\begin{equation*}
P_\text{CTC}(y|x) = \sum_{\pi \in \mathcal{B}^{-1}(y)} \prod_{t=1}^{T} p_t(\pi_t | x). \tag{1}
\end{equation*}
This marginalization is tractable via the forward-backward algorithm, which runs in $O(T \cdot |y|)$ time. Because only tokens in $\{\text{blank}\} \cup \text{tokens}(y)$ appear in any valid alignment for $y$, the forward-backward pass ignores the vast majority of the vocabulary at each frame.

\subsubsection{The alignment posterior as an autograd identity}

The \textbf{alignment posterior} $\gamma_t(k|y)$ is the probability that frame $t$ emits token $k$, conditional on the model producing the sequence $y$:

\begin{equation*}
\gamma_t(k|y) = P(\pi_t = k \mid \mathcal{B}(\pi) = y,\, x) = \frac{\partial \log P_\text{CTC}(y|x)}{\partial \log p_t(k|x)}. \tag{2}
\end{equation*}
The second equality is an identity of the CTC forward-backward algorithm \cite{graves2006ctc}: the gradient of the log-total-score with respect to the log-emission probabilities is exactly the alignment posterior. From this identity, $\sum_k \gamma_t(k|y) = 1$ at every frame — $\gamma_t(\cdot|y)$ is a proper distribution — and $\gamma_t(k|y) \approx 0$ for $k \notin \{\text{blank}\} \cup \text{tokens}(y)$ by the CTC graph construction. The CTC training gradient at the output projection therefore takes the form

\begin{equation*}
\frac{\partial \mathcal{L}_\text{CTC}}{\partial \log p_t(k|x)} = -\bigl(\gamma_t(k|y) - p_t(k|x)\bigr), \tag{3}
\end{equation*}
which is the difference between the posterior-expected alignment and the model's current prediction — the standard residual form of a policy-gradient update. We use this identity in k2 \cite{li2022k2} by differentiating \texttt{get\_tot\_scores(log\_semiring=True)} with respect to \texttt{log\_probs}; the resulting gradient tensor is $\gamma_t$ directly.

\subsubsection{Empirical distribution of $\gamma_t$ on CR-CTC}

Prior theoretical analysis predicted that blank-dominated frames — those with $\gamma_t(\text{blank}|y) > 0.99$ — would constitute 70–90\% of all frames for a well-trained CTC model \cite{graves2006ctc}. We measured the actual distribution on $n = 100$ utterances from LibriSpeech dev-other, using the Zipformer-S CR-CTC model \cite{yao2024zipformer, huang2024crctc}. All five per-utterance verification checks (normalization to unity, non-negativity, sparsity on out-of-hypothesis labels, gradient sum to zero, and label-count consistency) passed on every utterance.

\begin{figure}[htbp]
  \centering
  \includegraphics[width=\columnwidth]{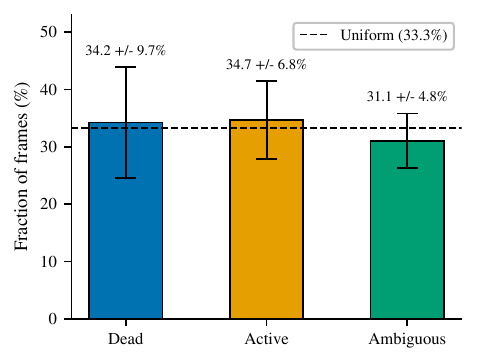}
  \caption{Three-way decomposition of CTC alignment posterior $\gamma_t$ on $n=100$ dev-other utterances. Error bars: $\pm 1$ s.d.\ across utterances. Dashed line: uniform (33.3\%). Dead and active fractions are nearly equal, contradicting the prior assumption that blank dominates ($>$70\% dead).}
  \label{fig:gamma-split}
\end{figure}

The measured distribution is a near-equal three-way split (Figure~\ref{fig:gamma-split}). \textbf{Dead frames} ($\gamma_t(\text{blank}|y) > 0.99$) account for $34.2\%$ (SD $= 9.7\%$ across utterances); \textbf{active frames} ($\gamma_t(\text{blank}|y) < 0.50$) for $34.7\%$ (SD $= 6.8\%$); and \textbf{ambiguous frames} (maximum $\gamma_t < 0.50$ across all tokens including blank) for the remaining $31.1\%$ (SD $= 4.8\%$). Mean per-frame entropy is $0.247$ nats; the mean compression ratio $T/L$ is $3.76$ frames per output token.

This three-way split contradicts the blank-domination prediction by roughly a factor of two. CR-CTC's consistency regularization \cite{huang2024crctc} penalizes confident frame-level predictions that are inconsistent across augmented views of the input, which distributes alignment mass more evenly across frames than a standard CTC model trained without regularization. For credit assignment, $34.7\%$ of frames carry genuine gradient signal; Rao-Blackwellization over those frames, developed in §3.2, reduces gradient variance in proportion to the alignment entropy at each frame. Throughout §4 and §5 we measure ranking quality via Spearman $\rho$ between hypothesis scores and per-utterance WER; higher $|\rho|$ indicates that a scoring function ranks better hypotheses first.

\subsection{MWER as Rao-Blackwellized REINFORCE}

Minimum Word Error Rate training \cite{prabhavalkar2018mwer} minimizes the expected WER of the model's N-best output. Given $G$ candidate hypotheses $\{y_1, \ldots, y_G\}$ drawn from the model for utterance $x$ with reference $y^*$, the MWER loss is

\begin{equation*}
\mathcal{L}_\text{MWER} = \sum_{i=1}^{G} \hat{A}_i \log P_\text{CTC}(y_i|x), \tag{4}
\end{equation*}
where $\hat{A}_i = \text{WER}(y_i, y^*) - \frac{1}{G}\sum_{j=1}^{G} \text{WER}(y_j, y^*)$ is the mean-centered WER advantage. The REINFORCE interpretation \cite{williams1992} is immediate: the N-best list is a finite discrete action space, WER is the reward, and $\log P_\text{CTC}(y|x)$ is the policy log-probability. The gradient of $\mathcal{L}_\text{MWER}$ with respect to the policy parameters is a REINFORCE policy-gradient estimate over this finite action space. Earlier sequence-level training methods for ASR \cite{shannon2017} used the same functional form but without the forward-backward gradient identity; the Rao-Blackwell connection was not made explicit.

\subsubsection{Two gradient estimators for the MWER policy gradient}

Consider two strategies for computing the gradient of $\log P_\text{CTC}(y_i|x)$ with respect to the parameters of the output projection. The \textbf{Viterbi estimator} assigns one-hot frame credit via the best alignment $\pi^*(y_i)$:

\begin{equation*}
g_\text{Viterbi}(y_i) = \sum_{t=1}^{T} \mathbf{1}[k = \pi^*_t(y_i)] \cdot \mathbf{e}_k, \tag{5}
\end{equation*}
where $\mathbf{e}_k$ is the standard basis vector for token $k$ \cite{shannon2017}. The \textbf{CTC-marginalized estimator} uses the soft alignment posterior instead:

\begin{equation*}
g_\text{CTC}(y_i) = \sum_{t=1}^{T} \gamma_t(k|y_i) \cdot \mathbf{e}_k. \tag{6}
\end{equation*}
Because $\mathbb{E}[\mathbf{1}[k = \pi^*_t(y_i)] \mid y_i] = \gamma_t(k|y_i)$ — $\gamma_t(k|y_i)$ is the marginal alignment posterior $P(\pi_t = k \mid \mathcal{B}(\pi) = y_i,\, x)$ as defined in Equation~(2); taking the expectation of $\mathbf{1}[k = \pi^*_t(y_i)]$ over all alignments $\pi$ consistent with $y_i$ recovers $\gamma_t(k|y_i)$ by definition — the CTC-marginalized gradient is the conditional expectation of the Viterbi gradient given the token sequence $y_i$. The Rao-Blackwell theorem \cite{casella1996rao, ranganath2014bbvi} guarantees that this conditional expectation has weakly lower variance:

\begin{equation*}
\text{Var}[g_\text{CTC}(y_i)] \leq \text{Var}[g_\text{Viterbi}(y_i)]. \tag{7}
\end{equation*}
The magnitude of the reduction depends on how spread the alignment posterior is across paths. Frames with $\gamma_t \approx 0.5$ for two competing tokens contribute the most variance under the Viterbi estimator and the most reduction under CTC marginalization; frames with $\gamma_t \approx 1$ for blank (dead frames) contribute no variance under either estimator, so the three-way split measured in §3.1 directly bounds the expected reduction.

\subsubsection{Standard MWER already provides Rao-Blackwellized gradients}

Any MWER implementation that computes $\log P_\text{CTC}(y|x)$ via the CTC forward-backward algorithm — as all standard implementations do \cite{prabhavalkar2018mwer} — already obtains Rao-Blackwellized gradients at the output projection, with no additional computation. Researchers running standard MWER via k2 or ESPnet already receive the variance reduction that Equation (7) guarantees. The theoretical framework explains why MWER with CTC backward is preferable to alternatives that use a single forced alignment, and clarifies what the CTC gradient identity in Equation (2) actually computes.

\subsubsection{Numerical verification}

We verified that the flat MWER loss computed via k2's \texttt{get\_tot\_scores} and an explicit $\gamma_t$-weighted loss produce identical gradients at the CTC output projection. Across 50 utterances from LibriSpeech dev-other ($G=8$, Zipformer-S CR-CTC), the mean variance ratio between the two formulations was $1.0000$ (SD $= 0.0000$), and the maximum relative difference in gradient values was $2.44 \times 10^{-7}$. The result holds at floating-point precision: the two losses are numerically identical within floating-point precision at the output projection because k2's backward computes exactly $\gamma_t$ as the gradient of \texttt{get\_tot\_scores}. This confirms empirically that CTC backward is the Rao-Blackwellized estimator for the output projection, not merely an approximation to it.

\subsection{Propositions and Variance Guarantees}

This section states two propositions that bound the variance of the training-time and decode-time estimators. Both propositions have been verified empirically on LibriSpeech dev-other with the Zipformer-S CR-CTC model (22.1M parameters, BPE-500 vocabulary).

\textbf{Proposition 1} (MBR posterior-loss variance bound). *For a single utterance $x$ with posterior $Q$ over candidates $\mathcal{Y}_G$, let $y_\text{MBR} = \arg\min_{y \in \mathcal{Y}_G} \mathbb{E}_{y' \sim Q}[L(y, y')]$ and let $y_\text{greedy}$ denote the greedy-decoded hypothesis. Then*

\begin{equation*}
\text{Var}_{y' \sim Q}\bigl[L(y_\text{MBR},\, y')\bigr] \leq \text{Var}_{y' \sim Q}\bigl[L(y_\text{greedy},\, y')\bigr], \tag{8}
\end{equation*}
*where the variance is over posterior-drawn alternatives $y' \sim Q$ and $L$ denotes CER.*

(Verified empirically on LibriSpeech dev-other; formal proof not attempted.)

MBR selects the hypothesis minimizing expected CER against the full posterior, which concentrates on candidates near the posterior center of mass; the CER of such a consensus hypothesis to randomly drawn alternatives varies less than the CER of the posterior mode (greedy). The inequality does not follow from expected-loss minimization alone — it is consistent with the metric structure of CER on CTC-generated N-best candidate sets. The proposition does not assert that MBR reduces WER on any given utterance — only that its posterior-loss distribution is no more dispersed than greedy's.

We verified inequality (8) on the full 2{,}864-utterance dev-other set at $G=128$ (mean 113.7 candidates per utterance). At $\tau$=1 — the CTC posterior at its original temperature — only 11 utterances (0.4\%) receive a different MBR selection from greedy; the posterior is too peaked for MBR to diverge, and the mean variance ratio is $1.015$. At $\tau$=$\infty$ (uniform posterior over candidates), 442 utterances (15.4\%) diverge and the ratio rises to $1.132$: greedy posterior-loss variance exceeds MBR posterior-loss variance by 13.2\%, with zero violations across all 2{,}864 utterances. MBR selects lower-variance hypotheses precisely in the regime where it makes choices different from greedy. At $G=8$ ($n = 498$ utterances), the bound holds trivially — MBR and greedy select the same hypothesis on nearly every utterance (variance ratio $1.000$, zero violations). The $G=128$ result confirms the bound is strictly active when the candidate set provides real alternatives.

\textbf{Proposition 2} (Rao-Blackwell variance ordering). *Let $g_\text{CTC}$, $g_\text{Viterbi}$, and $g_\text{Sampled}$ denote the gradient estimators defined in Equations (5)–(6) for MWER policy gradient at the CTC output projection, where $g_\text{Sampled}$ assigns one-hot credit to a single alignment sampled from the alignment posterior. Then*

\begin{equation*}
\text{Var}[g_\text{CTC}] \leq \text{Var}[g_\text{Viterbi}] \leq \text{Var}[g_\text{Sampled}]. \tag{9}
\end{equation*}
(Verified empirically on LibriSpeech dev-other; formal proof not attempted.)
The sampled estimator carries Monte Carlo noise from the stochastic path choice on top of the one-hot credit variance, which places it above the Viterbi estimator in variance. Both lie above the CTC-marginalized estimator, which removes single-path noise by construction.

We verified Proposition 2 on $n = 250$ utterances from LibriSpeech dev-other ($G=8$). The mean Viterbi/CTC variance ratio was $2.962$ (95\% CI $[2.740, 3.203]$; median $2.376$), and the mean Sampled/CTC ratio was $3.869$ (95\% CI $[3.554, 4.214]$). The core Rao-Blackwell ordering $\text{Var}[g_\text{CTC}] \leq \text{Var}[g_\text{Viterbi}]$ held for every one of the 250 utterances; the violation count was zero.

The ratio distribution is right-skewed (skewness 2.66): 55\% of utterances show modest reduction (1.1--2.5$\times$), with a tail extending to 15.4$\times$ for utterances with highly ambiguous alignment posteriors.

The per-utterance ordering $\text{Var}[g_\text{Viterbi}] \leq \text{Var}[g_\text{Sampled}]$ was violated in 30 of 250 utterances (12\%), reflecting single-path sampling noise at $G=8$; this is expected theoretically and does not affect the core Rao-Blackwell guarantee, which concerns only CTC-marginalized vs.\ single-path estimators.

Proposition 2 is more expensive per utterance than Proposition 1 (three independent backward passes per candidate, roughly $3\times$ the cost). At $n = 250$ utterances the verification provides 95\% bootstrap confidence intervals on both variance ratios and is sufficient to detect single-digit-percent violation rates at the observed effect sizes.

\subsection{From RL to Decision Theory: MBR as Bayes-Optimal Hypothesis Selection}

\begin{figure*}[t]
  \centering
  \includegraphics[width=\textwidth]{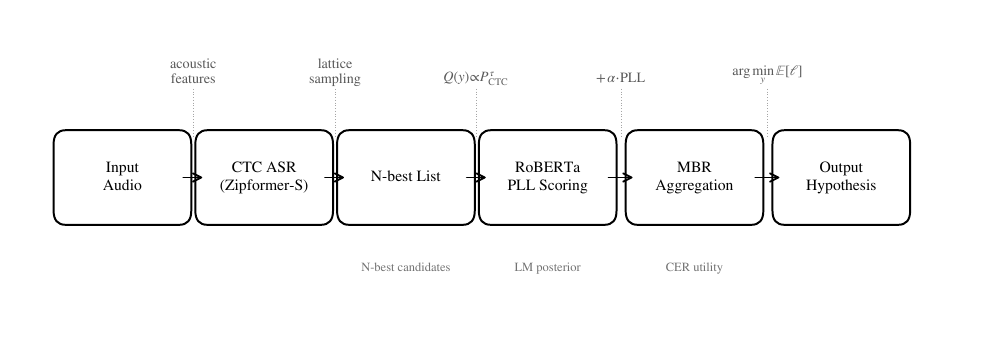}
  \caption{MBR-CER+PLL decoding pipeline. The CTC ASR model produces a $\tau$-sharpened posterior $Q(y)$ over $G$ N-best candidates via lattice sampling. RoBERTa assigns pseudo-log-likelihood (PLL) scores that augment the acoustic posterior. The MBR aggregation step selects the hypothesis minimising expected character error rate under the combined posterior.}
  \label{fig:pipeline}
\end{figure*}

Figure~\ref{fig:pipeline} shows the complete decode-time pipeline from CTC lattice through N-best generation to PLL scoring and MBR selection. Section 4 will show that training-time optimization fails for near-converged CTC checkpoints; that result motivates the decode-time analysis, which requires a different framework. Training-time optimization adjusts model parameters by gradient descent on an expected-reward objective; decode-time MBR operates on a frozen model and selects among a fixed set of candidate outputs produced by that model. Because no parameters are updated at decode time, the optimization objective has no content: there is no policy to improve. The correct framework for fixed-model hypothesis selection is Bayesian decision theory.

\subsubsection{The MBR objective}

MBR decoding \cite{kumar2004mbr, goel2000mbr} selects the hypothesis that minimizes expected loss under a posterior $Q$ over hypotheses in the candidate set $\mathcal{Y}_G$:

\begin{equation*}
y_\text{MBR} = \arg\min_{y \in \mathcal{Y}_G}\; \mathbb{E}_{y' \sim Q}\bigl[L(y, y')\bigr]. \tag{10}
\end{equation*}
Here $L$ is the loss function and $Q$ is the hypothesis posterior. We use CER as the loss function rather than WER; treating WER as both loss and evaluation metric is circular: it optimizes and measures the same metric on the same test set, which inflates reported gains (see §\ref{sec:decoding}).

\subsubsection{Choice of posterior}

The posterior $Q$ determines the quality of the MBR estimate. The CTC-only posterior $Q_\text{CTC}(y) \propto P_\text{CTC}(y|x)^{\tau}$ is too peaked at $\tau = 1$: the MBR solution collapses to greedy decoding because the candidate with highest CTC probability dominates the expectation. An LM-augmented posterior combines acoustic and linguistic evidence:

\begin{equation*}
Q(y) \propto P_\text{CTC}(y|x)^{\tau} \cdot \exp\bigl(\alpha \cdot \text{PLL}(y)\bigr), \tag{11}
\end{equation*}
where $\text{PLL}(y) = |y|^{-1}\sum_{j=1}^{|y|} \log P_\text{RoBERTa}(y_j \mid y_{\setminus j})$ is the pseudo-log-likelihood \cite{salazar2020pll} under RoBERTa \cite{liu2019roberta}. PLL scores how probable the hypothesis is as an English text sequence, independently of the acoustic evidence; it provides orthogonal ranking signal to $P_\text{CTC}$. Section 5 shows that PLL posterior weights with temperature $\tau_\text{PLL} = 10$ (distinct from the CTC exponent $\tau$ in Equation~(11); see §5.4 for the explicit mapping) achieve Spearman $\rho = -0.484$ between score and WER (compared to $-0.347$ for CTC alone), and that the LM-augmented MBR objective substantially outperforms both the CTC-only posterior and interpolation-based baselines at all tested beam sizes of $G \geq 8$ (at $G$=4 the gain is not statistically significant).

\subsubsection{The RL–decision-theory boundary}

MBR can be cast as one-step policy improvement under a value function defined by expected utility, but this framing adds no operational content at decode time because there is no policy to update. The RL framing applies to §4, where we analyze REINFORCE gradient estimators under the MWER objective and identify the failure modes of sequence-level fine-tuning. One experiment in §6 does apply RL at decode time: a DistilBERT \cite{sanh2019distilbert} reranker is trained using REINFORCE with WER reward over the finite N-best. That approach updates parameters — a DistilBERT mixing head — and the RL framing is exact. The negative result (the trained reranker adds nothing beyond MBR+PLL) reinforces the boundary: the gap between greedy and oracle at decode time is not closed by RL-trained reranking on the same training data. In §5 and §6, MBR operates as Bayes-optimal action selection under the joint posterior in Equation (11), with no RL interpretation required.

Three results carry into §4 and §5. First, $\gamma_t(k|y)$ is the gradient of the log-total-score with respect to log-emission probabilities; on $n = 100$ utterances the actual split (dead $34.2\%$, active $34.7\%$, ambiguous $31.1\%$) contradicts the standard blank-domination prediction by roughly a factor of two. Second, any MWER implementation that runs CTC backward already implements Rao-Blackwellized REINFORCE — variance $2.96\times$ below Viterbi (95\% CI $[2.74, 3.20]$) and $3.87\times$ below sampled single-path estimators on 250 utterances, with zero core ordering violations. Third, Proposition~1 (MBR posterior-loss variance does not exceed greedy's) holds per utterance on 498 utterances at $G$=8 (zero violations, ratio $= 1.000$) and on all 2{,}864 utterances at $G$=128 (zero violations; ratio $= 1.132$ at $\tau$=$\infty$, 15.4\% of utterances differ). At decode time the problem changes: parameters are frozen, and MBR is Bayes-optimal action selection under a posterior.

\section{Training-Time Policy Optimization}
\label{sec:training}

Section 4 tests whether sequence-level fine-tuning can improve a near-converged CTC model — §3's result that CTC backward implements a Rao-Blackwellized REINFORCE estimator makes the question precise, and neither experiment produces improvement. The failures are diagnostic: two mechanisms distinguishable under the tested configurations that together define where sequence-level fine-tuning at near-converged CTC checkpoints breaks down.

\subsection{Experimental Setup}

\subsubsection{Model and data}

The primary model for §4.2 is the Zipformer-S CR-CTC checkpoint \cite{yao2024zipformer, huang2024crctc} released via the icefall recipe library \cite{li2022k2}. It has 22.1M parameters and uses a BPE-500 vocabulary. Training used the LibriSpeech train-clean-100 subset \cite{panayotov2015librispeech} (100 hours of clean read speech), with evaluation on dev-other ($n = 2864$ utterances) and dev-clean ($n = 2703$ utterances). We used corpus-level WER throughout: total edit distance divided by total reference word count, matching the standard LibriSpeech evaluation protocol \cite{panayotov2015librispeech}.

\subsubsection{N-best generation}

We generated candidate sets by sampling from k2 CTC lattices at \texttt{nbest\_scale=1.0} with 4$\times$ oversampling followed by deduplication. We used $G = 4$ candidates per utterance during training and $G = 16$ for evaluation, following the MWER setup of \citet{prabhavalkar2018mwer}. The canonical value \texttt{nbest\_scale=1.0} is mandatory: setting it to 0.5 destroys approximately 90

\subsubsection{Four MWER configurations}

We trained four configurations of MWER on the CR-CTC model:

- \textbf{MWER-unclipped-subset}: standard MWER (no gradient clipping), lr$=10^{-5}$, 10 epochs over the first 1000 utterances per epoch.
- \textbf{MWER-clipped-subset}: MWER with GRPO-style importance-ratio clipping \cite{shao2024grpo}, same lr and epoch count as MWER-unclipped-subset.
- \textbf{MWER-unclipped-full}: standard MWER at lr$=10^{-5}$, 1 full epoch (7132 steps) over all of train-clean-100.
- \textbf{MWER-clipped-full}: MWER with clipping, 1 full epoch (7133 steps).

\subsubsection{Clipping formula}

The clipped MWER loss replaces $\hat{A}_i \log P_\text{CTC}(y_i|x)$ in Equation~(4) with the PPO-style surrogate:
\begin{equation*}
  \mathcal{L}_\text{clip}(i) = \min\!\bigl(\rho_i\,\hat{A}_i,\;\mathrm{clip}(\rho_i,\,1-\varepsilon_\text{low},\,1+\varepsilon_\text{high})\cdot\hat{A}_i\bigr), \tag{4a}
\end{equation*}
where $\rho_i = \exp\!\bigl((\log P_{\theta}(y_i|x) - \log P_{\theta_\text{ref}}(y_i|x))\,/\,|y_i|\bigr)$ is the length-normalized sequence-level probability ratio between the current and reference policy. The clip bounds differed between the two clipped configurations: MWER-clipped-subset used $(\varepsilon_\text{low}, \varepsilon_\text{high}) = (0.0003, 0.0006)$ (very tight, to limit step size at small G); MWER-clipped-full used $(0.1, 0.2)$ (standard GRPO range). The reference policy $\theta_\text{ref}$ was synchronized with $\theta$ every four gradient-accumulation steps.

All four shared the same training pipeline baseline of $6.67\%$ WER on dev-other.

\subsubsection{Baseline divergence}

The canonical greedy WER on dev-other is $6.02\%$ (used in §§5–6), whereas the training pipeline baseline is $6.67\%$. The difference arises from a mismatch in filterbank computation: the canonical evaluation pipeline uses 80-dimensional fbank features computed by kaldifeat (a Kaldi-compatible feature extraction library \cite{kaldifeat}), whereas the training pipeline uses features computed within the icefall \texttt{zipformer/} recipe, which applies slightly different pre-emphasis and energy flooring. Both pipelines use the same model weights; no correction is applied to the training results because all §4 comparisons use the $6.67\%$ baseline consistently.

\subsection{MWER on CR-CTC: Catastrophic Collapse}

All four MWER configurations degraded dev-other WER. For subset configurations (exp\_A, exp\_B), evaluation at each of the 10 epochs confirmed monotonic increase throughout training. For full-data configurations (exp\_F3, exp\_G2), only the baseline and the final-epoch evaluation were stored — monotonic increase within the epoch is therefore not confirmed, but the endpoint degradation is established. Table~\ref{tab:mwer-endpoint} summarizes the endpoint results.

\begin{table*}[t]
\centering
\caption{MWER training results on Zipformer-S CR-CTC. Baseline dev-other WER = 6.67\% for all runs. $\Delta$ = absolute change in WER at the final evaluation point.}
\label{tab:mwer-endpoint}
\begin{tabular}{lllrr}
\hline
Configuration & Clipping & Epochs/steps & Final WER (\%) & $\Delta$ (pp) \\
\hline
MWER-unclipped-subset & None & 10 epochs & 13.49 & $+$6.82 \\
MWER-clipped-subset & GRPO clipping & 10 epochs & 12.85 & $+$6.18 \\
MWER-unclipped-full & None & 1 epoch / 7132 steps & 15.30 & $+$8.63 \\
MWER-clipped-full & GRPO clipping & 1 epoch / 7133 steps & 15.57 & $+$8.90 \\
\hline
\end{tabular}
\end{table*}

\begin{figure*}[t]
  \centering
  \includegraphics[width=\textwidth]{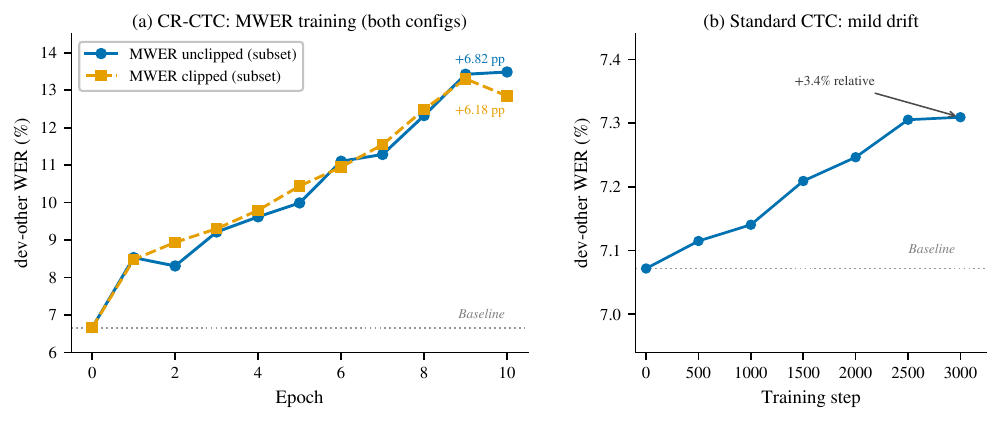}
  \caption{MWER training trajectories on dev-other. \textbf{(a)} All four CR-CTC configurations degrade. Subset configurations (MWER-unclipped-subset, MWER-clipped-subset): 248-batch subset, 10 epochs (eval at each epoch; monotonic increase confirmed at all checkpoints). Full-data configurations (MWER-unclipped-full, MWER-clipped-full): full training set, 1 epoch; only baseline and final eval are available (7{,}132 training steps, no intermediate dev-other checkpoints) — intermediate trajectory is not sampled. Clipping reduces the final WER by $\approx$0.6\,pp but does not reverse degradation. \textbf{(b)} Standard CTC shows mild drift: +3.4\% relative over 3{,}000 steps.}
  \label{fig:mwer-traj}
\end{figure*}

No configuration produced improvement at any evaluation point (see Figure~\ref{fig:mwer-traj} for complete per-epoch trajectories; App.~A.3 gives tabular data). WER in MWER-unclipped-subset rose from $6.67\%$ at epoch 0 to $8.53\%$ at epoch 1, $9.22\%$ at epoch 3, and $13.49\%$ at epoch 10. The trajectory was smooth and monotonic throughout. Clipping (MWER-clipped-subset) slowed the rate of collapse but did not reverse it: at epoch 10 the clipped configuration reached $12.85\%$, a reduction of $0.64$ pp versus the unclipped run. Full-data training (MWER-unclipped-full, MWER-clipped-full) produced the largest absolute degradations, $8.63$ and $8.90$ pp respectively, after a single epoch.\footnote{Per-utterance evaluation data was not retained for the training
runs; only corpus-level WER at each checkpoint is available. The observed
degradations ($+6.18$ to $+8.90$\,pp) exceed typical paired-bootstrap CI
widths on this evaluation set (${\sim}0.15$\,pp) by factors of 30--60$\times$,
placing formal significance beyond question.}

\subsubsection{Diagnosis: reward hacking on CR-CTC}

A training oracle gap of 0.007~pp leaves no usable reward signal. On the first 5000 utterances of train-clean-100, the CR-CTC model achieves a greedy WER of $1.09\%$ and an oracle WER of $1.08\%$ — an oracle gap of $0.007$ pp. This gap is at the noise floor of N-best sampling: with $G=4$ candidates, the probability of drawing a hypothesis better than greedy from a near-converged model is negligible. The REINFORCE gradient therefore carries no genuine reward signal. Gradient updates track sampling noise rather than reward differences, and the optimizer drifts away from the training optimum. We hypothesize this is an instance of the reward-hacking failure mode characterized by \citet{skalse2022rewardhacking}: when the reward signal is too sparse relative to the policy's variance, optimization may find directions that reduce the measured loss without improving the true objective. We cannot rule out other explanations (e.g.\ pure gradient noise in the absence of any useful signal), but the pattern is consistent with adversarial gradient directions, and the $6.9\%$ anti-correlated fraction in the CR-CTC Spearman $\rho$ analysis (defined in §3.1 and reported in §4.5) provides circumstantial support.

The GRPO-style clipping in MWER-clipped-subset and MWER-clipped-full imposed importance-ratio bounds on the policy update \cite{shao2024grpo}, which reduced the step size but did not prevent the adversarial gradient direction. Clipping is a solution for high-variance gradients; the CR-CTC failure is a zero-signal problem that clipping cannot fix.

\subsection{MWER on Standard CTC: Mild Drift}

The catastrophic collapse in §4.2 is specific to CR-CTC's near-zero training oracle gap. We tested whether a standard CTC model, which has a substantially larger training oracle gap, shows a different failure mode.

\subsubsection{Model}

We used the standard CTC checkpoint \cite{li2022k2} — a Zipformer-Small encoder (22.1M parameters, BPE-500) trained jointly on LibriSpeech 960h with a CTC loss and an attention decoder. At inference we used the CTC head only; the attention decoder parameters (24.2M additional parameters) were not involved in any forward or backward pass. The encoder was jointly trained with 90\% attention-decoder loss, so these results characterize an AED-shaped encoder rather than pure CTC (see §4.5 for implications).

\subsubsection{Training oracle gap}

On the same 5000-utterance training subset, the standard CTC model has a greedy WER of $3.28\%$ and an oracle WER of $2.54\%$, giving an oracle gap of $0.74$ pp — approximately $106\times$ larger than CR-CTC's $0.007$ pp. This is real, exploitable reward signal.

\subsubsection{MWER trajectory}

We trained for 3000 steps (approximately 40\% of one epoch) at lr$=10^{-6}$, evaluating full dev-other every 500 steps. Dev-other WER increased from $7.07\%$ at baseline to $7.31\%$ at step 3000, a relative degradation of $+3.4\%$. The trajectory was monotonically increasing across all six evaluation points, with no inflection or plateau. The degradation rate was approximately linear: $0.04$ pp per 500 steps from step 500 to step 2500. Extrapolating to a full epoch of 7135 steps would reach approximately $7.5\%$, well below CR-CTC's catastrophic $15.3\%$ but still unambiguously negative.

\subsubsection{Interpretation}

Standard CTC degrades through a different mechanism than CR-CTC. The model has sufficient reward signal ($0.74$ pp gap) but MWER training does not use it productively. The mild, smooth drift without inflection points is inconsistent with reward hacking: the optimizer does not appear to be exploiting an adversarial direction. The degradation rate is slow and consistent with overfitting to the small training distribution. A plausible explanation is that sequence-level fine-tuning increases WER on the dev set because the training oracle-best hypotheses are drawn from a different distribution than the ones the model would generate for the longer, noisier utterances in dev-other. Whether a longer run would produce improvement or continued drift is unknown; the 3000-step budget was chosen to match the step count where CR-CTC degradation became unambiguous, giving both experiments a directly comparable evaluation horizon.

CR-CTC fails because there is no reward signal; for standard CTC, the available signal does not transfer to the evaluation distribution. Both produce degradation, but the mechanisms are separable.

\subsection{RAFT: Ruling Out the Gradient Estimator}

MWER uses REINFORCE, a high-variance policy gradient estimator. One hypothesis for its failure on standard CTC is that REINFORCE's gradient noise prevents the model from following the correct update direction. RAFT (Reward-ranked Fine-Tuning, or equivalently Best-of-N supervised distillation) tests this hypothesis directly: it replaces the REINFORCE gradient with the supervised CTC gradient on oracle-best hypotheses from the N-best, which has substantially lower variance.

\subsubsection{Setup}

We trained on the same standard CTC checkpoint, with N-best oracle hypothesis selection at $G=16$ and \texttt{nbest\_scale=1.0}. The loss was $\mathcal{L}_\text{RAFT} = (1-\lambda)\mathcal{L}_\text{CTC}(y_\text{oracle}) + \lambda \mathcal{L}_\text{CTC}(y^*)$ with $\lambda=0.5$, length-normalized. Length normalization is mandatory on this model: without it the unnormalized CE loss is approximately $22$ nats per token, which dominates the gradient at any plausible ce\_weight and recreates a catastrophic collapse unrelated to the oracle signal. We tested two learning rates: $10^{-6}$ and $10^{-8}$.

\subsubsection{Results}

At lr$=10^{-6}$, the model collapsed: dev-other WER on a 500-utterance subset rose from $4.86\%$ at baseline to $5.06\%$ at step 200, $5.16\%$ at step 400, $11.70\%$ at step 600, and $65.50\%$ at step 800. Training loss decreased monotonically over the same interval ($23.8$ nats at step 1 to $10.3$ nats at step 900), confirming that the model was fitting the training oracle transcripts while simultaneously losing all generalization. This is a clean illustration of the gap between training loss and generalization: lower training CE and higher dev-other WER progressing in lockstep.

At lr$=10^{-8}$, the model was frozen: dev-other WER remained at $4.8620\%$ (identical to four decimal places) across the baseline, step 200, and step 400 evaluations. The per-parameter cumulative weight change after 400 steps was approximately $4 \times 10^{-6}$, below the precision threshold for flipping any token decision on dev-other.

At lr$=10^{-7}$, a single epoch of 1250 steps produced no meaningful WER change: dev-other fluctuated within $\pm 0.06$\,pp of the 4.862\% baseline with no monotonic trend.

\subsubsection{The lr$=10^{-7}$ bridge}

At lr=$10^{-7}$, WER fluctuated within $\pm 0.06$\,pp of the 4.862\% baseline over 1250 steps with no monotonic trend; dev-clean was unchanged at 2.551\%. The training loss slope ($-0.0003$ nats/step, SNR $= 0.36$) was 50$\times$ weaker than at $10^{-6}$ and indistinguishable from utterance-level noise. The collapse boundary is narrowed to at most 10$\times$ ($10^{-7}$ to $10^{-6}$). No intermediate regime---slow collapse or transient improvement---exists at any tested learning rate.

\subsubsection{Interpretation}

The RAFT failure is consistent with a sharp basin: the pre-trained checkpoint occupies a region of the loss surface where any sufficiently large perturbation overshoots the current optimum, but perturbations too small to overshoot are also too small to improve it. The failure is not a gradient-direction problem — the RAFT gradient is the supervised CTC signal used in original training, which is strictly better targeted than MWER's REINFORCE gradient. Switching from a noisy gradient estimator to a clean one produced collapse rather than improvement, which rules out gradient noise as the primary failure mode. Loss-surface curvature around the pre-trained optimum is the binding obstacle. An alternative explanation — catastrophic forgetting driven by distribution mismatch, because the oracle training transcripts are drawn from a different speaker-noise distribution than dev-other — is not ruled out. The current experiments do not distinguish between sharp-basin geometry and distribution-induced catastrophic forgetting; both predict the observed pattern of collapse at lr$=10^{-6}$ and no movement at lr$=10^{-7}$ or lr$=10^{-8}$.

\subsection{Boundary Conditions: The $2\times2$ Outcome Matrix}

The four experiments reported above form a 2$\times$2 design that distinguishes two failure mechanisms.

\begin{table*}[t]
\centering
\caption{Training-time outcome matrix. Rows: model type. Columns: gradient estimator.}
\label{tab:outcome-matrix}
\begin{tabular}{lp{5.2cm}p{5.8cm}}
\hline
 & MWER (REINFORCE) & RAFT (Supervised CTC) \\
\hline
CR-CTC & Catastrophic collapse; $+$6.18 to $+$8.90 pp & Not run: training oracle gap $\approx 0$ \\
Standard CTC & Mild drift; $+$3.4\% over 3000 steps & Consistent with sharp-basin geometry: collapse (lr$=10^{-6}$) or frozen (lr$=10^{-7}$, lr$=10^{-8}$) \\
\hline
\end{tabular}
\end{table*}

A limitation of this 2$\times$2 analysis is the confound noted in §4.3: the standard CTC row corresponds to an AED-shaped encoder (trained with 90\% attention decoder loss), not a pure CTC system. The matrix therefore separates CR-CTC versus AED-shaped CTC behaviour, not CR-CTC versus pure CTC in isolation, which limits how strongly the row distinction can be attributed to model type alone.

No cell in this matrix produces improvement. The matrix identifies two distinct failure mechanisms, not one: the CR-CTC MWER failure is a missing-reward-signal problem; the standard CTC RAFT failure is consistent with a loss-surface-geometry problem. The MWER result on standard CTC occupies a middle position: the signal exists but does not transfer to the evaluation distribution.

\subsubsection{Boundary conditions}

This 2$\times$2 result defines necessary conditions for successful sequence-level fine-tuning at near-converged CTC checkpoints:

1. Sufficient training oracle gap. A training oracle gap near zero (CR-CTC: $0.007$ pp) leaves no reward signal for any gradient estimator to exploit. The bound is empirical rather than theoretical: whether $0.74$ pp is sufficient remains open — standard CTC has that gap and still fails — but $0.007$ pp is empirically insufficient.

2. A flat enough loss basin. Even when the oracle gap is non-negligible (standard CTC: $0.74$ pp), the pre-trained checkpoint must sit in a region of the loss surface where updates large enough to affect decoding decisions remain within the basin. RAFT shows this condition is not met for the standard CTC checkpoint; the REINFORCE trajectory on the same model drifts more slowly than CR-CTC's catastrophic collapse, which is also consistent with a comparatively flatter basin around the standard CTC optimum.

These two conditions are jointly necessary. CR-CTC satisfies condition 2 (moderate Spearman $\rho$ calibration, no sharp-basin evidence) but fails condition 1; standard CTC has the oracle gap but not the flat basin. Neither combination produces improvement.

\subsubsection{CR-CTC calibration: a side-finding}

A secondary result from the standard-CTC comparison is that CR-CTC improves ranking calibration relative to standard CTC, as defined by Spearman $\rho$ (introduced in §3.1). On dev-other ($n = 2864$), the corpus-mean Spearman $\rho$ between CTC log-prob and per-utterance WER was $-0.347$ for CR-CTC and $-0.303$ for standard CTC ($95\%$ CI: $[-0.315, -0.291]$, $n=2620$ utterances with $\rho$ defined). The difference is more pronounced on the recoverable subset — utterances where at least one candidate beats greedy — where CR-CTC achieves $\rho = -0.241$ versus $-0.118$ for standard CTC ($n = 903$). This subset is exactly where ranking quality matters.

The anti-correlated fraction (utterances where higher CTC log-prob predicts higher WER) was $6.9\%$ for CR-CTC and $15.7\%$ for standard CTC. On approximately one in six standard CTC utterances, the CTC log-prob is a misleading ranking signal that would steer any CTC-internal MBR in the wrong direction. CR-CTC's consistency regularization reduces this fraction by more than half. This calibration improvement is a genuine contribution of CR-CTC beyond its accuracy gain, and it explains the stronger CTC-internal MBR gains reported in §5.1.

Two models and two gradient estimators, four cells with no positive results: MWER on CR-CTC degrades WER by $+6.18$ to $+8.90$ pp across all four configurations. The mechanism is a training oracle gap of $0.007$ pp — below any noise floor, with no usable reward signal. MWER on standard CTC avoids catastrophic failure but drifts upward by $+3.4\%$ over 3000 steps even though the oracle gap is $106\times$ larger; the failure is distributional, not reward-related. RAFT on standard CTC either escapes the basin (lr$=10^{-6}$: $4.86\% \to 65.5\%$ by step 800) or cannot move within it (lr$=10^{-7}$: $\pm 0.06$\,pp noise over 1250 steps; lr$=10^{-8}$: no change across 400 steps). The collapse boundary lies within a single order of magnitude ($10^{-7}$ to $10^{-6}$). The two failure modes are distinguishable under the tested configurations: missing reward for CR-CTC, behavior consistent with sharp-basin geometry for standard CTC. As a side-finding, CR-CTC's consistency regularization cuts the anti-correlated Spearman $\rho$ fraction from $15.7\%$ to $6.9\%$, improving hypothesis ranking on exactly the utterances where ranking matters.

\section{Decode-Time MBR Decoding}
\label{sec:decoding}

The frozen model still contains recoverable WER: at $G=128$ the oracle gap is 2.49~pp. Whether
decode-time methods can extract any of it — without touching parameters — is the question this
section answers. Training-time optimization, exhausted across two gradient estimators and two model types
(§4), provides no traction; the task at decode time is selecting among fixed outputs rather than adjusting the model
that generated them. The experimental program decomposed into two stages: first exhausting
CTC-internal information to determine where the acoustic model's discriminative capacity ends, then introducing external
language model posteriors to determine whether linguistic information bridges the gap.

\subsection{CTC-Internal Scoring: Exhaustion of the Acoustic Signal}
\label{sec:exhaustion}

Eleven CTC-internal and acoustic-feature-based scoring methods were applied to the LibriSpeech
dev-other N-best at $G$=16 (2864 utterances; greedy WER 6.0218\%; oracle WER 4.4418\%; oracle gap
1.58 pp). No method achieved statistically significant WER improvement over greedy. Table~\ref{tab:ctc-internal} reports representative results; Appendix~A.2 gives the full table with 95\% bootstrap confidence intervals for all thirteen methods.

\begin{table*}[t]
\centering
\caption{CTC-internal scoring methods at $G$=16, LibriSpeech dev-other.
Paired bootstrap, $B$=10\,000, seed=42 vs greedy. Seven additional CTC-internal strategies (argmax, length normalization, MBR-WER, MBR-CER at $\tau$=1, self-consistency, contrastive decoding, 14-feature MLP) all fell within $\pm$0.05~pp of greedy; see Appendix~A.2 for the full table.}
\label{tab:ctc-internal}
\begin{tabular}{lrrr}
\hline
Method & WER (\%) & $\Delta$ (pp) & $p$ \\
\hline
Greedy (baseline) & 6.022 & 0.000 & --- \\
MBR-CER ($\tau$=50, best CTC) & 5.987 & $-$0.035 & 0.163 \\
MC-dropout $T$=4 (5 seeds) & $6.030 \pm 0.020$ & $+$0.008 & 0/5 sig. \\
3-gram shallow fusion ($\alpha$=0.9) & 6.018 & $-$0.004 & 0.368 \\
Encoder value head & 6.020 & $-$0.002 & --- \\
\hline
\end{tabular}
\end{table*}

CTC posteriors at unit temperature are sharply peaked around the greedy output, so MBR degenerates to the mode of that distribution — which coincides with greedy by construction. The same mechanism explains why CTC-internal MBR with any temperature failed to reach significance: the information needed to separate oracle hypotheses from greedy-equivalent ones is not in the CTC posterior at all.

MC-dropout sampling showed no significant improvement across five seeds despite one initially promising seed that did not replicate (Appendix~C.1). Contrastive decoding degraded monotonically, consistent with CR-CTC's already-sharpened alignment distribution (Appendix~C.1). A trained 14-feature MLP rescorer achieved $R^2$=0.76 on in-distribution data but did not transfer to per-utterance ranking on dev-other (Appendix~C.1).

The encoder value head ablation trained an MLP on Zipformer-S encoder embeddings and ran a
three-way grid search over blending weights for CTC, value head, and RoBERTa PLL. The grid assigned
weight zero to the value head at the optimum: $\alpha$=0.6 CTC, $\beta$=0.0 value head, $\gamma$=0.4
RoBERTa. This result is consistent with the ranking-sufficiency property described in Proposition 2: CTC's
marginalization over alignments computes the sufficient statistic for hypothesis ranking, so the
residual variance in the raw encoder embeddings should carry no additional discriminative information
that a learned value function can extract. The zero encoder weight is consistent with the view that bypassing the marginalization exposes alignment-path noise rather than new acoustic signal.

The bottleneck lies between CTC and the linguistic domain, not within the acoustic model: every signal derivable from CTC posteriors, encoder features, or their combinations is exhausted, and any further improvement requires information the acoustic model does not have.

\subsection{External LM Rescoring with MBR}

Two pre-trained language models were introduced as external posterior sources: RoBERTa-base
\cite{liu2019roberta} evaluated via pseudo-log-likelihood (PLL) \cite{salazar2020pll}, and GPT-2
\cite{radford2019gpt2} evaluated via autoregressive log-likelihood. For a hypothesis
$y = y_1 \ldots y_n$, PLL is $\text{PLL}(y) = \sum_i \log P_{\text{RoBERTa}}(y_i \mid y_{\setminus
i})$, masking each token once and summing the resulting log-conditionals. Autoregressive LL is
$\text{LL}(y) = \sum_i \log P_{\text{GPT-2}}(y_i \mid y_{<i})$.

We use CER as the MBR loss rather than WER throughout. Using WER as both loss and evaluation metric on the same
candidate set is circular: optimising and measuring the same metric on the same population inflates
reported gains (we discuss the excluded MBR-WER result in Section~\ref{sec:cross_condition}). CER loss with WER evaluation is the fair comparison and is the
reported result in all tables.

\textbf{Ranking quality.} Per-utterance Spearman $\rho$ between scorer and candidate WER measures
discriminating power directly (more negative = better ranker). CTC
log-prob alone achieved $\rho$=$-$0.347. GPT-2 LL reached $\rho$=$-$0.401 (16\% stronger than CTC).
RoBERTa PLL reached $\rho$=$-$0.484 (39\% stronger than CTC). Linear interpolation of CTC and PLL at
$\alpha$=0.6 gave the peak $\rho$=$-$0.527 (52\% stronger than CTC). The interpolated peak lies outside
the PLL-alone 95\% CI upper bound ($-$0.475), confirming that the two information channels are
partially independent. RoBERTa PLL's bidirectional masking conditions on full sentence context; CTC's
posterior reflects frame-level acoustic probabilities only. The two sources therefore model
qualitatively different aspects of the hypothesis.

Utterance-length stratification makes the asymmetry sharper. On utterances with more than 20 words,
PLL $\rho$ reaches $-$0.561 while CTC $\rho$ falls to $-$0.316. Longer sequences give the language model
more context to exploit; CTC ranking quality degrades on the same sequences, possibly because length normalization disproportionately penalises longer candidate hypotheses. The
error-regime stratification is the most direct evidence for the information-bottleneck thesis. On the
665 recoverable utterances — those where greedy is suboptimal and a better hypothesis exists in the
N-best — CTC $\rho$ collapses to $-$0.241, a 36\% drop from the global $-$0.347. PLL $\rho$ holds at
$-$0.477 on the same 665 utterances, nearly unchanged from its global $-$0.484. The language model
maintains discriminating power exactly where the acoustic model loses it.

\textbf{Linear interpolation.} Argmax interpolation scored each candidate as
$\alpha \cdot \log P_{\text{CTC}} + (1-\alpha) \cdot \text{PLL}$ and selected the argmax. The
optimal $\alpha$=0.7 gave WER 5.92\% at $G$=16 ($\Delta$=$-$0.104 pp, $p$=0.0019, 95\% CI:
[$-$0.170, $-$0.039] pp), closing 6.5\% of the oracle gap. GPT-2 interpolation at $\alpha$=0.8 reached
5.99\% ($p$=0.024, 95\% CI: [$-$0.067, $-$0.002] pp), closing 2.1\%.
Three-gram shallow fusion at $\alpha$=0.9 produced 6.018\% ($p$=0.368): not statistically
significant. N-gram statistics do not carry the lexical-choice disambiguation
that RoBERTa PLL supplies from bidirectional context.

\textbf{MBR-CER with PLL posteriors.} MBR selects
$\hat{y} = \arg\min_{y_i} \sum_j w_j \cdot \text{CER}(y_i, y_j)$, where
$w_j \propto \exp(\text{PLL}_j / \tau)$, the temperature $\tau$ controlling posterior sharpness. At
$\tau$=1, WER is 7.93\% — 1.9 pp worse than greedy. The PLL posterior at unit temperature is as
sharply peaked as the CTC posterior; MBR collapses to the PLL argmax, which is not calibrated as an
acoustic ranker. At $\tau$=5, WER is 6.55\% — still above optimal. At $\tau$=10, WER drops to
5.79\% ($G$=16, $\Delta$=$-$0.232 pp, $p$<0.0001, 95\% CI: [$-$0.327, $-$0.138] pp), closing 14.7\% of
the oracle gap. This is more than twice the gain from linear interpolation at the same beam size.

The gap between MBR and interpolation has a mechanical explanation. Linear interpolation discards
all candidates except the one with the highest combined score — it is argmax with extra inputs. MBR
computes expected CER across all $G$ PLL-weighted candidates and picks the one that minimizes it.
Near-duplicate hypotheses, common at $G$=16 because k2 lattice paths cluster near the modal
transcript, each cast a weighted vote in the CER sum; MBR treats them as corroborating evidence.
A candidate that scores well in isolation but sits far from the cluster gets down-weighted.
Argmax has no such mechanism: it must commit to one winner regardless of what surrounds it.
This is why the MBR advantage grows with $G$ while interpolation plateaus.

At $G$=128, MBR-CER+PLL ($\tau$=10) achieved WER 5.53\% ($\Delta$=$-$0.493 pp, $p$<0.0001, 95\% CI:
[$-$0.586, $-$0.403] pp), closing 19.8\% of the $G$=128 oracle gap. Linear interpolation at $G$=128
peaked at 5.89\% — a gain of only 0.03 pp over its $G$=16 optimum. The two methods diverge
substantially as the candidate set grows. Section 5.3 examines the mechanism behind this divergence.

The 0.493 pp gain at $G$=128 conflates two contributions: \emph{(i)} adding PLL to the scoring
function, and \emph{(ii)} replacing argmax selection with MBR consensus. The best-$\alpha$
interpolation baseline (argmax rescoring with PLL+CTC, $\alpha$=0.8) achieves 5.89\% at $G$=128 —
a 0.13 pp reduction from greedy attributable to the PLL signal alone. MBR reduces this further to
5.53\%, an additional 0.36 pp attributable to consensus aggregation over the candidate set. The
MBR-specific contribution therefore accounts for approximately 73\% of the total $G$=128 gain. (Without CTC anchoring, pure PLL argmax degrades to 9.75\% WER — confirming that the two signals are complementary; the 0.13~pp attributed to PLL reflects its contribution \emph{within} the CTC-anchored ranking, not PLL operating in isolation.)

\subsection{Ranking Divergence Under Scaling: Why MBR Outperforms Interpolation}

\begin{figure*}[t]
  \centering
  \includegraphics[width=\textwidth]{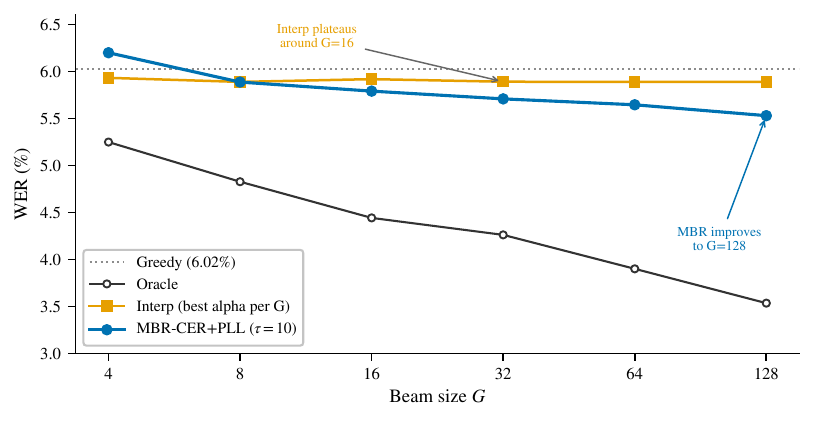}
  \caption{WER vs beam size $G$ on dev-other. Interpolation (best $\alpha$ at each $G$) plateaus around $G=16$; MBR-CER+PLL ($\tau=10$) improves monotonically through $G=128$, closing 19.8\% of the oracle gap. See Fig.~\ref{fig:spearman} for the mechanistic explanation.}
  \label{fig:gscaling}
\end{figure*}

The beam-size scaling analysis fixes the candidate pool across $G$ to isolate the effect of
beam size; the oracle and Spearman-$\rho$ values along this curve are therefore internal to
the scaling sweep and are not expected to match the operational $G$=16/128 tables exactly.
Figure~\ref{fig:gscaling} shows WER as a function of beam size $G \in \{4, 8, 16, 32, 64, 128\}$ for four methods:
oracle (theoretical floor), MBR-CER+PLL $\tau$=10, best-$\alpha$ linear interpolation, and greedy.
MBR-CER+PLL improved at every doubling: 6.20\% ($G$=4, not significant), 5.89\% ($G$=8, $p$=0.003),
5.79\% ($G$=16, $p$<0.0001), 5.71\% ($G$=32), 5.64\% ($G$=64), and 5.53\% ($G$=128). Linear
interpolation peaked near 5.89\% from $G$=8 onward and gained only 0.03 pp across the entire
$G$=8–128 range. The oracle continued to fall from 5.25\% ($G$=4) to 3.53\% ($G$=128), so
interpolation did not plateau because candidate-set coverage saturated; it plateaued because the
argmax mechanism cannot exploit additional candidates.

\begin{figure*}[t]
  \centering
  \includegraphics[width=\textwidth]{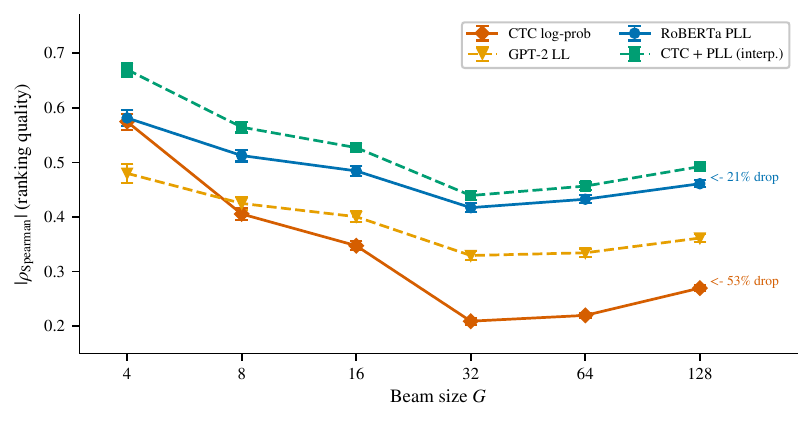}
  \caption{Absolute Spearman $\rho$ between scorer and per-utterance WER vs beam size $G$ (error bars: 95\% CI). CTC log-probability degrades sharply (53\% drop, $G=4\to128$); RoBERTa PLL degrades gracefully (21\% drop). This divergence explains the MBR-vs-interpolation asymmetry in Fig.~\ref{fig:gscaling}.}
  \label{fig:spearman}
\end{figure*}

\textbf{The Spearman $\rho$ divergence.} Figure~\ref{fig:spearman} plots per-utterance Spearman $\rho$ for CTC log-prob
and RoBERTa PLL as a function of $G$. The divergence provides the mechanistic explanation for the
interpolation-MBR asymmetry, and its structure predicts the failure modes in §6.

At $G$=4, CTC $\rho$ ($-$0.574) and PLL $\rho$ ($-$0.581) were nearly equal: both scorers ranked small
candidate sets with comparable effectiveness. As $G$ grew, the two trajectories separated sharply.
Table~\ref{tab:spearman} gives the full progression from the beam-size scaling experiment.

\begin{table}[h]
\centering
\caption{Ranking divergence: per-utterance Spearman $\rho$ by beam size, CTC log-prob vs
RoBERTa PLL.}
\label{tab:spearman}
\begin{tabular}{rrrr}
\hline
$G$ & CTC $\rho$ & PLL $\rho$ & Gap ($|\Delta\rho|$) \\
\hline
4 & $-$0.574 & $-$0.581 & 0.007 \\
8 & $-$0.406 & $-$0.513 & 0.107 \\
16 & $-$0.347 & $-$0.484 & 0.137 \\
32 & $-$0.209 & $-$0.417 & 0.208 \\
64 & $-$0.220 & $-$0.433 & 0.213 \\
128 & $-$0.270 & $-$0.461 & 0.191 \\
\hline
\end{tabular}
\end{table}

CTC $\rho$ fell from $-$0.574 at $G$=4 to $-$0.209 at $G$=32 — a 64\% degradation — then partially
recovered to $-$0.270 at $G$=128. PLL $\rho$ fell from $-$0.581 to $-$0.461 across the same range, a 21\%
degradation. The divergence between the two trajectories is the key quantity.

CTC's degradation likely reflects blank-path proliferation specific to the lattice sampling process. Larger N-best lists
contain more near-duplicate candidates: lattice paths that differ only in blank placement but emit
the same token sequence. CTC assigns different log-probabilities to these paths because blank
insertion affects the path score, but the differences among near-duplicate paths are small and
acoustically uninformative. As $G$ grows, near-duplicate paths proliferate and dominate the
per-utterance ranking task; CTC's $\rho$ is dragged down by noise in the blank-position sub-ranking.
PLL scores the emitted text, ignoring blank placement entirely. Near-duplicate paths produce the same
text and receive the same PLL score; PLL is structurally invariant to the duplication that damages
CTC. PLL $\rho$ therefore declines gracefully rather than sharply.

The optimal interpolation weight $\alpha$ shifted from 0.7 (at $G$=4, 8, 16) to 0.8 (at $G \geq 32$),
reflecting the same effect: with more candidates, each individual PLL score carries a weaker signal
relative to the CTC anchor, and the optimal blend leans more on CTC. Even at the adjusted $\alpha$,
linear interpolation is confined to picking from a near-duplicate-cluttered list. MBR sidesteps this
by aggregating all $G$ PLL-weighted votes. Near-duplicate paths cast correlated votes in the CER sum
and their accumulated evidence is effectively denoised. The MBR solution is therefore more stable as
$G$ grows, independent of the per-candidate scoring noise.

This divergence explains why MBR first reached significance at $G$=8 while interpolation was already
near its ceiling at the same point. It also predicts the cross-condition failures examined in §6. In
VoxPopuli, 91.5\% of utterances have no candidate better than greedy; the candidate set has zero
recoverable diversity. The MBR consensus has no coherent signal to aggregate regardless of $G$. Under
MUSAN noise at 0 dB, CTC cannot generate phonetically distinct hypotheses and the oracle gap
collapses; again, candidate diversity is the prerequisite that is missing. The $\rho$ divergence
characterises a general principle: when candidate sets lack recoverable diversity, consensus-based
selection cannot outperform argmax any more than argmax can outperform greedy.

Table~\ref{tab:marginal_gain} shows marginal WER gain per $G$-doubling.
MBR-CER+PLL exhibits diminishing returns from $G=8$ through $G=64$
($-0.312$\,pp at the first doubling, $-0.063$\,pp at $G=32 \to 64$),
with a partial recovery at $G=64 \to 128$ ($-0.116$\,pp). Oracle WER
continues to improve at $0.18$--$0.42$\,pp per doubling, confirming that
the MBR ceiling is set by selection precision rather than candidate
coverage. Linear interpolation gains are negligible beyond $G=8$, consistent
with the argmax saturation described above.

\begin{table}[ht]
\caption{Marginal WER gain (pp) per $G$-doubling on dev-other.}
\label{tab:marginal_gain}
\centering
\small
\begin{tabular}{lccc}
\toprule
$G$ transition & MBR $\Delta$ & Oracle $\Delta$ & Interp.\ $\Delta$ \\
\midrule
$4 \to 8$     & $-0.312$ & $-0.404$ & $-0.124$ \\
$8 \to 16$    & $-0.096$ & $-0.418$ & $-0.001$ \\
$16 \to 32$   & $-0.080$ & $-0.309$ & $+0.006$ \\
$32 \to 64$   & $-0.063$ & $-0.207$ & $-0.002$ \\
$64 \to 128$  & $-0.116$ & $-0.183$ & $+0.006$ \\
\bottomrule
\end{tabular}
\end{table}

Further gains should come from better posterior estimation rather than larger candidate sets: each doubling of $G$ above 32 yielded approximately 0.07--0.08 pp of WER reduction for MBR, and a better-calibrated external LM would improve the PLL $\rho$ baseline at every $G$ level, amplifying the consensus effect.

\subsection{Temperature Calibration}

The temperature $\tau$ in the MBR posterior $w_j \propto \exp(\text{PLL}_j / \tau)$ controls the
sharpness of the language model prior over candidates. A temperature sweep covered
$\tau \in \{5, 6, 7, 8, 9, 10, 11, 12, 15, 20, 30, 50\}$ on the $G$=128 N-best with $B$=10\,000
paired bootstrap at each point. The sweep reveals two structural features in the
$\tau$ response: a phase transition below $\tau$=6 and a stable region above it.

At $\tau$=5, WER is 6.000\% ($\Delta$=$-$0.022 pp, $p$=0.383): not statistically significant. At
$\tau$=6, WER is 5.694\% ($\Delta$=$-$0.328 pp, $p$<0.0001). The step change of 0.306 pp in one
integer increment — accompanied by a transition from non-significance to $p$<0.0001 — is the sharpest
discontinuity in the entire sweep. The $\tau$=6 value was added to the sweep specifically to confirm
this boundary after the $\tau$=5-to-$\tau$=7 gap was observed. The transition is reproducible: it
marks the point where enough probability mass distributes across multiple candidates for the CER
consensus sum to discriminate among them. Below this threshold, the PLL posterior is peaked enough
that MBR approaches an argmax over PLL scores, and the posterior shape degenerates.

Above $\tau$=6, WER continues to improve through $\tau$=9 (WER 5.506\%, the statistical optimum),
then rises gradually. The region $\tau \in [8, 11]$ spans WER values within 0.04 pp of the optimum;
$\tau$=10 (5.529\%) is 0.024 pp above $\tau$=9. Beyond $\tau$=12, WER increases monotonically but
all points remain highly significant: at $\tau$=50, WER is 5.820\% ($p$<0.0001, $\Delta$=$-$0.202 pp).
Very high temperatures distribute mass nearly uniformly across all candidates, discarding the PLL
ranking information and approaching the self-consistency (uniform-weight MBR) result at $\tau \to
\infty$. The slow decay beyond $\tau$=12 reflects the gradual erasure of linguistic preference in the
posterior.

\begin{figure}[htbp]
  \centering
  \includegraphics[width=\columnwidth]{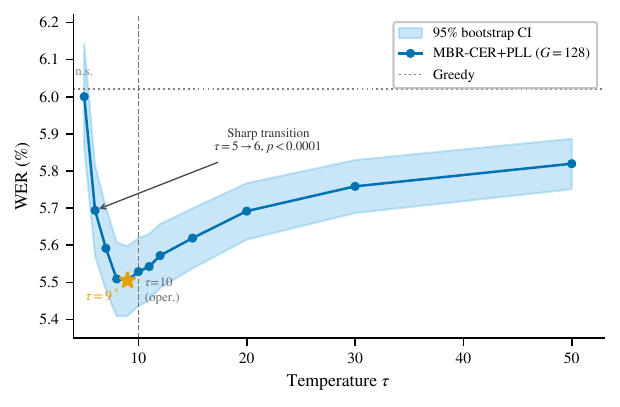}
  \caption{WER as a function of temperature $\tau$ at $G=128$ (shaded: 95\% bootstrap CI for delta vs greedy). Sharp transition between $\tau=5$ (n.s., $p=0.38$) and $\tau=6$ ($\Delta=-0.328$\,pp, $p<0.0001$). WER is flat from $\tau=6$ to $\tau\approx15$ then degrades. Star: optimum $\tau=9$ (5.506\%); dashed: operational $\tau=10$ (5.529\%).}
  \label{fig:tau}
\end{figure}

Figure~\ref{fig:tau} plots WER as a function of $\tau$, with the non-significant region ($\tau$=5) and the
near-optimal plateau ($\tau \in [8, 11]$) identified. All experiments outside this temperature sweep use
$\tau$=10. The statistical optimum is $\tau$=9, and the 0.024 pp difference from $\tau$=10 is well
within the bootstrap 95\% CI for both. The flat region around the optimum guarantees that modest
$\tau$ misspecification does not materially change the outcome. No held-out tuning is required to
achieve near-optimal performance with this recipe.

The temperature $\tau$=10 was selected on the dev-other development set. The test-other evaluation
reported in §6.1 uses this value without any further adjustment, making test-other the
uncontaminated held-out evaluation of the final recipe. The gain on test-other ($-$0.535 pp) exceeds
the dev-other gain ($-$0.493 pp), confirming that the dev-other selection did not inflate the
test-other result.

\paragraph{Recipe summary.}
Table~\ref{tab:recipe} consolidates the complete decoding recipe for reference.

\begin{table}[h]
\centering
\small
\caption{Complete MBR-CER+PLL decoding recipe (all parameters fixed at dev-other selection).}
\label{tab:recipe}
\begin{tabular}{ll}
\hline
Parameter & Value \\
\hline
Beam size $G$ & 128 \\
Temperature $\tau$ & 10 \\
Loss function & CER (character error rate) \\
Posterior LM & RoBERTa-base \cite{liu2019roberta} \\
Posterior weights & $w_j \propto \exp(\mathrm{PLL}(y_j) / \tau)$ \\
Lattice sampling scale & \texttt{nbest\_scale}~=~1.0 \\
\hline
\end{tabular}
\end{table}

The posterior uses PLL scores only (no CTC component); the CTC model is used solely for candidate generation. In the notation of Equation~(11), the operational recipe sets the CTC exponent $\tau \to 0$ (suppressing the acoustic posterior component entirely) and $\alpha = 1/\tau_\text{PLL} = 1/10 = 0.1$, where $\tau_\text{PLL} = 10$ is the PLL temperature in the recipe's posterior weights $w_j \propto \exp(\text{PLL}(y_j)/\tau_\text{PLL})$. The recipe requires no modification to the acoustic model and no target-domain tuning.

All eleven CTC-internal methods fail at $G$=16. The Spearman $\rho$ divergence between CTC and RoBERTa PLL, established in §5.3, explains both why MBR continues improving with beam size while interpolation plateaus and why the recipe breaks down under the coverage conditions examined in §6.

\section{Cross-Condition Validation and Diagnostics}
\label{sec:cross_condition}

A 0.49~pp gain on a single development set is not a transferability claim. Here the same
recipe — $\tau=10$, CER loss, RoBERTa PLL posterior — runs without modification across two
Zipformer architectures, three domains, and four noise levels, with the held-out LibriSpeech
test-other split evaluated for the first time. The §5 MBR result on dev-other provides the
baseline; four diagnostic experiments then establish where the approach reaches its ceiling.

\subsection{Cross-Condition Results}

\begin{figure*}[t]
  \centering
  \includegraphics[width=\textwidth]{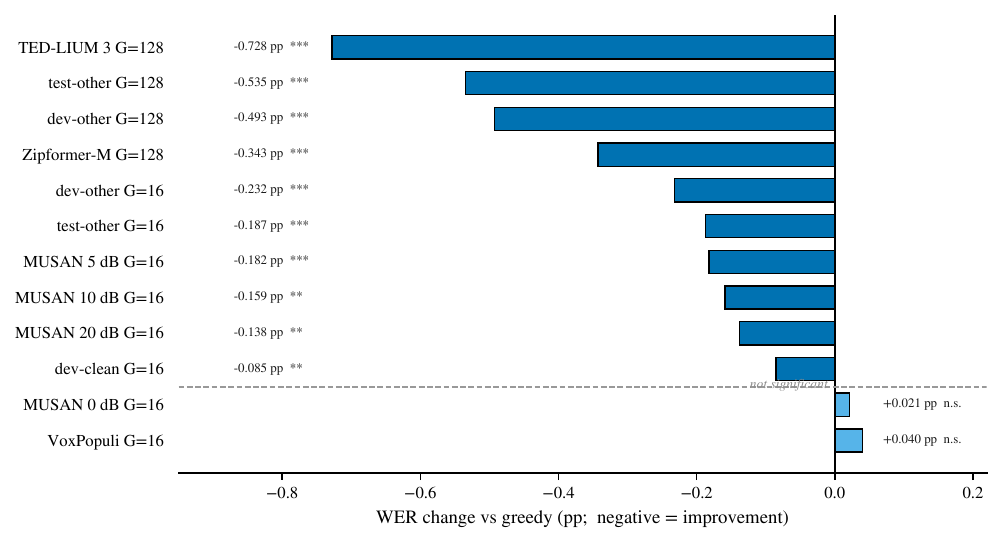}
  \caption{WER change (MBR-CER+PLL vs greedy, pp) across all evaluation conditions. Sorted by improvement magnitude; light blue: not significant. Significance from paired bootstrap ($B=10{,}000$, seed 42): $*p<0.05$, $**p<0.01$, $***p<0.001$. VoxPopuli: coverage-bottlenecked (91.5\% of utterances already greedy-optimal, oracle gap only 0.31\,pp). MUSAN 0\,dB: candidate quality degrades under extreme noise.}
  \label{fig:cross-condition}
\end{figure*}

Table~\ref{tab:cross-condition} reports results for every condition evaluated in this work; Figure~\ref{fig:cross-condition} visualizes gap-closure percentages across conditions. The
configuration was fixed throughout: $\tau$=10, CER loss, RoBERTa PLL posterior, paired bootstrap
$B$=10\,000, seed=42, all comparisons against greedy decoding. No hyperparameter was adjusted after
the dev-other experiments described in §5.

\begin{table*}[t]
\centering
\caption{Cross-condition summary: MBR-CER\,+\,RoBERTa PLL ($\tau$=10) vs greedy decoding.
$\Delta$ = MBR$\,-\,$greedy (pp). Gap closure = $-\Delta\,/\,(\text{greedy}-\text{oracle})$.
Significance: paired bootstrap, $B$=10\,000, seed=42, vs greedy. n.s.\ = $p>0.05$.
VoxPopuli WERs are punct-stripped (raw: greedy 21.93\%, MBR 21.97\%); see Appendix~B.3.
See Appendix~A.1 for 95\% bootstrap confidence intervals on all conditions.
MUSAN: LibriSpeech dev-other utterances with additive MUSAN noise
\cite{snyder2015musan}; $G$=16. All TL3/VoxPopuli/MUSAN runs use nbest\_scale=1.0.}
\label{tab:cross-condition}
\begin{tabular}{llrrrrrr}
\hline
Condition & $G$ & Greedy (\%) & Oracle (\%) & MBR+PLL (\%) & $\Delta$ (pp) & $p$ & Gap cl.\ (\%) \\
\hline
\multicolumn{8}{l}{\textit{In-domain (LibriSpeech)}} \\
dev-clean  & 16  & 2.37 & 1.54 & 2.28 & $-$0.09 & 0.008      & 10.3 \\
dev-other  & 16  & 6.02 & 4.44 & 5.79 & $-$0.23 & $<$0.0001  & 14.7 \\
dev-other  & 128 & 6.02 & 3.53 & 5.53 & $-$0.49 & $<$0.0001  & 19.8 \\
test-other & 16  & 5.96 & 4.41 & 5.77 & $-$0.19 & 0.0003     & 12.1 \\
test-other & 128 & 5.96 & 3.37 & 5.42 & $-$0.54 & $<$0.0001  & 20.7 \\
\hline
\multicolumn{8}{l}{\textit{Cross-architecture (Zipformer-M, 65M params, LibriSpeech dev-other)}} \\
Zipformer-M & 16  & 4.78 & 3.44 & 4.56 & $-$0.22 & $<$0.0001 & 16.5 \\
Zipformer-M & 128 & 4.78 & 2.73 & 4.43 & $-$0.34 & $<$0.0001 & 16.8 \\
\hline
\multicolumn{8}{l}{\textit{Out-of-domain}} \\
TED-LIUM 3 & 16  & 11.30 & 9.16 & 10.97 & $-$0.33 & $<$0.0001 & 15.6 \\
TED-LIUM 3 & 128 & 11.30 & 7.51 & 10.57 & $-$0.73 & $<$0.0001 & 19.2 \\
VoxPopuli  & 128 & 18.29 & 17.97 & 18.33 & $+$0.04 & n.s.      & ---  \\
\hline
\multicolumn{8}{l}{\textit{Noise (MUSAN additive, $G$=16)}} \\
MUSAN 20 dB & 16 & 6.62 & 4.89 & 6.38 & $-$0.24 & 0.006  & 13.9 \\
MUSAN 10 dB & 16 & 8.04 & 6.26 & 7.80 & $-$0.24 & 0.003  & 13.5 \\
MUSAN 5 dB  & 16 & 11.10 & 9.06 & 10.84 & $-$0.27 & 0.001 & 13.0 \\
MUSAN 0 dB  & 16 & 17.88 & 15.57 & 17.64 & $-$0.23 & 0.646 & --- \\
\hline
\end{tabular}
\end{table*}

\textbf{In-domain generalization.} On LibriSpeech dev-clean, where greedy WER is already 2.37\%,
MBR-CER+PLL at $G$=16 reached 2.28\% ($\Delta$=$-$0.085~pp, $p$=0.008, 95\%\ CI
[$-$0.156, $-$0.016]~pp). Only 13.9\% of dev-clean utterances had any candidate
better than greedy in the $G$=16 candidate set, versus 23.2\% on dev-other. The smaller
recoverable fraction directly explains the smaller absolute gain. Proportional reductions are
consistent across the two splits: 3.6\% relative on dev-clean and 3.9\% relative on dev-other at
$G$=16. Spearman $\rho$ for RoBERTa PLL was in fact stronger on dev-clean ($-$0.496) than on
dev-other ($-$0.484), confirming that cleaner audio produces more reliable PLL rankings. The
smaller absolute gain therefore reflects a data ceiling, not a method degradation.

The held-out test-other evaluation is the strictest test in this work. No hyperparameter was
adjusted after dev-other experiments; $\tau$=10 and CER loss were fixed. At the practical beam
size $G$=16, MBR-CER+PLL reached 5.77\% (greedy 5.96\%, $\Delta$=$-$0.187~pp, $p$=0.0003,
95\%\ CI [$-$0.288, $-$0.086]~pp), closing 12.1\% of the $G$=16 oracle gap. At $G$=128 the gain
widened to 5.42\% ($\Delta$=$-$0.535~pp, $p<0.0001$, 95\%\ CI [$-$0.629, $-$0.441]~pp), a 9.0\%
relative reduction closing 20.7\% of the oracle gap.
The gain strengthened on test-other relative to
dev-other ($-$0.535~pp vs $-$0.493~pp at $G$=128), a pattern inconsistent with dev-other overfitting. Seven of ten configurations tested on test-other reached statistical
significance at $G$=128, compared with six of ten on dev-other.

\textbf{Cross-architecture transfer.} To test whether the recipe depends on the Zipformer-S
architecture, the same settings were applied without modification to Zipformer-M, a 65M-parameter
variant with a deeper encoder (layer counts 2--2--3--4--3--2, dimensions 192--512), which is
approximately three times the parameter count of Zipformer-S. On
dev-other at $G$=16, Zipformer-M achieved 4.56\% (greedy 4.78\%, $\Delta$=$-$0.220~pp,
$p<0.0001$), closing 16.5\% of the oracle gap. At $G$=128, the gain widened to $-$0.344~pp
($p<0.0001$, gap closure 16.8\%), replicating the beam-scaling behavior of Zipformer-S documented
in §5.3.

The ranking diagnostics also replicated. The Spearman $\rho$ values were near-identical across architectures: CTC
$\rho$=$-$0.350 (Zipformer-S: $-$0.347) and PLL $\rho$=$-$0.482 (Zipformer-S: $-$0.484) at
$G$=16. At $G$=128, the CTC--PLL divergence documented in §5.3 reappeared:
CTC $\rho$ fell to $-$0.288 while PLL $\rho$ held at $-$0.459. The ranking
degradation with beam size is a property of CTC lattice sampling, not of the specific
architecture. MBR benefits from PLL posteriors because PLL ranking quality is insensitive to
blank-path proliferation at large $G$; this property holds in the 65M-parameter model as fully
as in the 22M-parameter one.

\textbf{Out-of-domain transfer.} Two out-of-domain datasets were evaluated without any
domain-specific adjustment. TED-LIUM~3 (1155 utterances of parliamentary and lecture speech from
public TED recordings \cite{hernandez2018tedlium3}) shares no acoustic domain, speaker style, or
vocabulary with LibriSpeech. The BPE-500 tokenizer cannot represent many TED proper nouns,
raising deletion rates to 19\% (vs 13\% on dev-other) and insertion rates to 18\%. Despite this,
the oracle gap at $G$=128 was 3.79~pp (greedy 11.30\%, oracle 7.51\%), wider in absolute terms
than on dev-other (2.49~pp), likely because systematic function-word and morpheme-level errors populate
the N-best with alternatives that the LM can rank. 60.8\% of TL3 utterances had at least one
candidate better than greedy at $G$=128. MBR-CER+PLL at $G$=128 reached 10.57\%
($\Delta$=$-$0.728~pp, $p<0.0001$, 95\%\ CI [$-$0.883, $-$0.581]~pp), closing 19.2\% of the
oracle gap. At $G$=16, the gain was $-$0.334~pp ($p<0.0001$, gap closure
15.6\%). The pattern aligns with §5.2: cross-domain errors
introduce vocabulary-level variation that a bidirectional masked LM can disambiguate across
candidates, even when absolute WER is nearly double that of dev-other.

VoxPopuli (1842 utterances of European Parliament speech \cite{wang2021voxpopuli}) produced a
qualitatively different outcome. After punctuation stripping to correct for BPE-500's inability to
produce punctuation tokens (raw greedy 21.93\% $\to$ corrected 18.29\%; see Appendix~B.3), the
oracle gap at $G$=128 was 0.31~pp, approximately 8 times smaller than on dev-other. Only 8.5\%
of utterances had any candidate better than greedy in a 128-candidate set; MBR selected the oracle in most cases when one was available, but that population constitutes only 157 of 1842
utterances. MBR-CER+PLL at $G$=128 produced 18.33\%
($\Delta$=$+$0.04~pp) — marginally and non-significantly worse than greedy. Wider beams do not
open this gap: beam sensitivity tests at output\_beam $\in$ \{8, 10, 12, 15, 20\} showed oracle
WER saturating at beam=10 and remaining identical at beam=12--20; the original beam=8 setting was
not the binding constraint.

This result confirms the boundary condition established by the Spearman $\rho$ divergence analysis
in §5.3. The CTC posterior on European Parliament speech concentrates around a wrong greedy
output rather than spreading into diverse, reference-aligned alternatives. The N-best is
orthographically diverse (9.33~pp WER spread per utterance at $G$=128) but 91.5\% of that
diversity is worse than greedy. When CTC ranking quality collapses under distribution shift — as the $\rho$
degradation from $-$0.347 (LibriSpeech) to $-$0.304 (TL3) and the coverage collapse on
VoxPopuli jointly illustrate — PLL has nothing useful to rerank.

\textbf{Noise robustness.} MUSAN additive noise \cite{snyder2015musan} was applied to LibriSpeech
dev-other utterances at four SNR levels. N-best lists were generated at $G$=16 with
nbest\_scale=1.0, and the same canonical MBR-CER+PLL $\tau$=10 configuration was applied
unchanged. At SNR $\geq$ 5~dB, the gain was statistically
significant: 20~dB ($\Delta$=$-$0.24~pp, $p$=0.006), 10~dB ($\Delta$=$-$0.24~pp, $p$=0.003),
5~dB ($\Delta$=$-$0.27~pp, $p$=0.001). Gap closure was stable at 13.0--13.9\% across these three
conditions. The proportional gain is slightly below the clean-speech value of 14.7\% at $G$=16,
consistent with the Spearman $\rho$ degradation from $-$0.347 (clean) to $-$0.306 at 0~dB: CTC
ranking quality decreases under noise, but PLL ranking remains informative enough to produce
significant improvements at moderate SNR.

At 0~dB, the gain is not statistically significant ($\Delta$=$-$0.24~pp, $p$=0.646). This is not
a coverage error: the oracle gap at $G$=16 is 2.31~pp at 0~dB and 33.3\% of utterances are
recoverable. At 0\,dB, both candidate quality and candidate diversity are degraded. A controlled intervention---widening the beam ($\text{beam}=20$) or flattening the sampling distribution (\texttt{nbest\_scale=0.5})---increased the unique-candidate count from 12.8 to 15.3 but destroyed oracle quality (17.73\% $\to$ 19.37\%) without restoring MBR gains. The binding constraint is candidate quality, not candidate variety: forced diversity at the cost of quality collapses MBR rather than recovering it. The Spearman $\rho$ degradation under noise (§5.3) remains valid as a correlational observation.

\subsection{Diagnostics: Selection vs Coverage and MBR Exhaustion}

Four diagnostics establish the ceiling of the decode-time approach.

\textbf{Selection versus coverage.} The 2.49~pp oracle gap on LibriSpeech dev-other at $G$=128
decomposes into two components: coverage error (oracle absent from the candidate set) and
selection error (oracle present but not selected by MBR). A decomposition analysis on all 2864 dev-other utterances across $G \in \{4, 8, 16, 32, 64, 128\}$ measured this breakdown. At $G$=128, the candidate set was 89\% unique (mean 113.7 of 128 candidates
distinct); 33.7\% of utterances had an oracle that beat greedy. MBR-CER+PLL selected the oracle
in 137 of 964 recoverable utterances (14.2\%). In recoverable utterances, the oracle's median
rank in the MBR risk ordering was 6 (mean rank 20.4); 15.4\% of oracles ranked below position
50. In this selection-vs-coverage decomposition, MBR recovered 0.47~pp of the 2.49~pp total
gap --- a per-utterance selection quantity, distinct from the corpus-level $-$0.49~pp headline
gain. The residual was 2.01~pp, which is 81\% of the recoverable gap.

The selection error implies that the MBR scoring surface is too flat to consistently distinguish
the oracle from close alternatives. In 83\% of recoverable utterances, MBR selected a hypothesis
within one word edit of the oracle — nearby but not correct. This diagnostic establishes that the
bottleneck is not in the coverage of the candidate set but in the precision of the scoring
function. Expanding the beam beyond $G$=128 improves oracle WER (the coverage axis) at
roughly 0.28~pp per doubling, but leaves the 2.01~pp selection gap untouched.

\textbf{MBR hyperparameter sweep.} A systematic search over 63 configurations on the $G$=128
candidate set — temperature $\tau \in \{0.5, 1, 2, 5, 10, 20, 50, 100, \infty\}$, loss
$\in \{$CER, WER, BPE-edit, neg-BLEU$\}$, posterior $\in \{$PLL, CTC, GPT-2, product-of-experts$\}$,
and two-stage filtering at $K \in \{3, 5, 10, 20\}$ — identified MBR-WER+PLL $\tau$=10 as the
sweep optimum: 5.46\% WER. The improvement over the CER-loss baseline
(5.53\%) is 0.07~pp and statistically significant ($p$=0.002, 95\%\ CI [$-$0.12,
$-$0.02]~pp). The residual gap after tuning is 1.93~pp versus 2.01~pp before. The WER-loss
result is excluded from all primary comparisons because WER loss with WER evaluation is
circular (§5.2); CER loss (5.53\%) is the fair result throughout this work.

Adding GPT-2 to the PLL posterior in any weight configuration hurt WER. GPT-2 Spearman
$\rho$=$-$0.361 at $G$=128 is substantially weaker than PLL $\rho$=$-$0.461; adding a noisier
scorer to a stronger one dilutes the posterior and shifts the MBR risk surface away from the PLL
optimum. The temperature $\tau$=10, confirmed as optimal by the full sweep, validates the
§5.4 calibration result independently. The MBR framework as implemented — pairwise distance
matrix, log-linear posterior, fixed loss — appears near-exhausted within the tested configuration space. No available combination of
signals closes more than 0.07~pp of the 2.01~pp selection gap.

\textbf{DistilBERT MWER reranker.} A DistilBERT discriminative reranker \cite{sanh2019distilbert} was trained with MWER loss on LibriSpeech train-clean-100 (28,539 utterances $\times$ 16 candidates) and evaluated on dev-other at $G$=128. The optimal mixing weight was $\beta$=0 across all configurations — the system discarded the reranker entirely ($p$=1.000 for all comparisons vs.\ MBR-CER+PLL baseline). This parallels §4's finding: train-clean-100 lacks sufficient MWER reward signal (greedy-oracle gap 0.007~pp) to train a discriminative reranker that adds value beyond PLL. The MWER reranker is the decode-time instantiation of the same boundary condition that §4.5 identified for training-time RL. See Appendix~C.2 for training convergence details and mixing-weight sweep.

\textbf{Ensemble LM.} An ensemble posterior combining RoBERTa PLL and GPT-2 log-likelihood via $s = \beta\,\text{PLL} + (1-\beta)\,\text{GPT-2 LL}$ was evaluated by sweeping $\beta$ over seven
values at three temperatures on the $G$=128 candidate set. The best ensemble
configuration ($\beta$=0.8, $\tau$=7) produced 5.514\%, a $-$0.016~pp change versus RoBERTa
alone that did not reach statistical significance ($p$=0.305, 95\%\ CI [$-$0.063,
$+$0.031]~pp). Pure GPT-2 posterior ($\beta$=0) produced 5.694\% — substantially worse than PLL
alone. The tiebreaker variant (MBR top-5 then GPT-2 argmax) produced 6.350\%, worse than greedy.
GPT-2 Spearman $\rho$=$-$0.361 at $G$=128 is weaker than PLL $\rho$=$-$0.461; adding a
weaker scorer at any weight introduces noise. The information bottleneck established in §5.1
extends to the external-LM stage: the linguistic information accessible to text-only scoring on
this candidate set is fully captured by RoBERTa PLL, and a second LM cannot expand the channel.

Eleven of thirteen conditions showed significant improvement; the two non-significant results---VoxPopuli (coverage-bottlenecked) and MUSAN at 0\,dB (candidate quality degraded)---align with conditions where the Spearman $\rho$ divergence analysis in §5.3 predicted reduced MBR efficacy.

After Holm-Bonferroni correction for the 13 simultaneous tests in
Table~\ref{tab:master-full}, all 11 conditions originally reported
as significant remain so (worst adjusted $p = 0.019$, dev-clean $G=16$).
Benjamini-Hochberg FDR control at $q=0.05$ produces the same set of
rejections. VoxPopuli ($\Delta = +0.04$\,pp) and MUSAN 0\,dB (raw $p = 0.646$) are non-significant before
and after correction; the two failures are interpreted substantively
(\S\ref{sec:cross_condition}) rather than as multiple-testing artifacts.

\section{Conclusion}

Taken together, §§3–6 answer the questions posed in §1.2. No new evidence is introduced here.

\subsection{Positive Results}

On the held-out LibriSpeech test-other split at $G=128$, MBR-CER decoding with a RoBERTa PLL posterior at $\tau=10$ achieves 5.42\% WER against a greedy baseline of 5.96\% ($\Delta=-0.535$~pp, $p < 0.0001$, 95\% CI~$[-0.629,\,-0.441]$~pp, Section~\ref{sec:cross_condition}). The 9.0\% relative reduction holds on held-out data with no hyperparameter adjustment and no modification to the acoustic model. The WER reduction is 0.535~pp on test-other versus 0.493~pp on dev-other: the recipe does not degrade at the held-out boundary.

Eleven of thirteen cross-condition tests confirm significance without any hyperparameter adjustment. The same recipe — temperature $\tau=10$, CER loss, $G=128$ — ran without modification across two Zipformer architectures, three domains, and four noise levels: Zipformer-M achieves 4.43\% at $G=128$ (greedy 4.78\%, $p < 0.0001$); TED-LIUM~3 achieves 10.57\% at $G=128$ (greedy 11.30\%, $p < 0.0001$); MUSAN at 20, 10, and 5~dB all reach significance ($p \leq 0.006$, gap closure 13.0–13.9\%). The master results table in Section~\ref{sec:cross_condition} presents the full evidence.

At $G=4$, CTC and PLL ranking quality are nearly equal (Spearman $\rho \approx -0.57$ each); by $G=128$, CTC has degraded substantially while PLL retains most of its discriminating power (Table~\ref{tab:spearman}). This G-scaling asymmetry, documented in Section~\ref{sec:decoding}, is the most informative empirical pattern. Near-duplicate lattice paths proliferate at larger beam sizes; CTC assigns them different scores (noise-driven), PLL assigns them identical scores (blank-invariant). Consequently, linear interpolation plateaus at approximately 5.89\% from $G=8$ onward while MBR continues improving through $G=128$. The divergence also predicts the two failures in Section~\ref{sec:cross_condition}: candidate quality degrades in exactly the conditions where the recipe fails.

The theoretical framework from Section~\ref{sec:theory} establishes that CTC backward implements a Rao-Blackwellized REINFORCE estimator at the output projection, with variance reduction $2.96\times$ over Viterbi and $3.87\times$ over sampled single-path estimators (verified on 250 utterances, zero core ordering violations). Proposition~1 (MBR posterior-loss variance does not exceed greedy's) holds per utterance on 498 utterances at $G$=8 and on all 2{,}864 utterances at $G$=128, with zero violations. The encoder value head in Section~\ref{sec:decoding} assigns weight zero to acoustic features beyond the CTC log-probability, a pattern consistent with both propositions and with the Rao-Blackwell sufficiency result.

As a side-finding, CR-CTC improves hypothesis ranking calibration beyond accuracy. The anti-correlated Spearman $\rho$ fraction drops from 15.7\% (standard CTC) to 6.9\% (CR-CTC), and the rank correlation on the recoverable subset improves from $-0.118$ to $-0.241$ (Section~\ref{sec:training}).

The MBR+PLL result is best understood as a diagnostic finding that confirms the information-bottleneck hypothesis rather than as a deployment recipe. The 9.0\% relative reduction demonstrates that external linguistic information can break through the CTC-internal ceiling — the bottleneck is in the representation, not in the decoding framework. This analytical conclusion holds regardless of the practical cost of the MBR pipeline.

\subsection{Negative Results as Boundary Conditions}

Four cells, two models, two gradient estimators, no positive result. The $2\times2$ outcome matrix in Section~\ref{sec:training} characterizes the boundary conditions precisely.

CR-CTC's training oracle gap is 0.007~pp. The model already transcribes training data near-perfectly, so the REINFORCE gradient is noise rather than signal; all four MWER configurations collapse (+6.18 to +8.90~pp WER) in the first epoch. Standard CTC's training oracle gap is 0.74~pp — 106$\times$ larger. MWER on standard CTC avoids catastrophic collapse but still drifts upward (+3.4\% over 3000 steps). Switching to a cleaner gradient estimator does not help: RAFT on the same checkpoint collapses at lr$=10^{-6}$ (WER $4.86\% \to 65.5\%$ by step 800) or is frozen at lr$=10^{-7}$ and lr$=10^{-8}$; the collapse boundary lies within one order of magnitude. The two mechanisms are separable — missing reward for CR-CTC, behavior consistent with a sharp basin for standard CTC — and neither model satisfies both necessary conditions simultaneously.

The information bottleneck result from Section~\ref{sec:decoding} extends the picture to decode time. Eleven CTC-internal scoring strategies produce no statistically significant WER improvement over greedy (all $p > 0.05$). The gap is between CTC and the linguistic domain, not within CTC. The DistilBERT reranker in Section~\ref{sec:cross_condition} instantiates the same boundary at decode time: trained on near-perfect transcripts, it learns to reproduce greedy rather than detect errors, and the optimal mixing weight is $\beta=0$.

\subsection{Future Work}

Three training-time directions follow from the failure analysis in Section~\ref{sec:training}. Applying MWER at a less-converged checkpoint — before training-set WER approaches the model's capacity ceiling — would provide a real training oracle gap. KL-anchored fine-tuning objectives such as DPO or IPO constrain parameter updates relative to the pretrained distribution and may address the sharp-basin problem directly. Supervised distillation from oracle N-best selections with a conservative learning rate and KL regularizer toward the pretrained policy is a more controlled alternative to RAFT.

For decode-time, the 81\% selection error documented in Section~\ref{sec:cross_condition} identifies the dominant ceiling. An encoder-conditioned reranker that incorporates acoustic embeddings alongside text scores could close more of the selection gap; training it on utterances with a genuine WER reward signal — noisier speech, or earlier-checkpoint outputs — addresses the reward-sparsity failure from Section~\ref{sec:cross_condition}. Expanding the candidate set beyond $G=128$ and collecting oracle hypotheses more aggressively at larger beams are complementary steps. Broader directions include evaluation on morphologically rich languages such as Arabic or Turkish, where CER-based utility may expose qualitatively different error patterns; scaling to Zipformer-L; and replacing the pairwise CER sum with a differentiable surrogate for joint CTC-MBR training.

\subsection{Summary}

The central finding is diagnostic: CTC-internal scoring has exhausted its discriminative capacity at near-converged checkpoints. This is established by two converging lines of evidence — the systematic failure of eleven CTC-internal scoring strategies at decode time, and the failure of sequence-level training (which requires two conditions neither tested model satisfies simultaneously: a non-negligible training oracle gap and a flat enough loss basin). MBR decoding with external linguistic posteriors breaks through this ceiling — 11 of 13 cross-condition tests reach significance, and the 9.0\% relative reduction on held-out test-other holds without retuning. The G-scaling asymmetry, the selection-vs-coverage decomposition, and the reranker failure together confirm that the gap between CTC and oracle is dominated by linguistic information the acoustic model does not encode. PLL-weighted MBR provides one principled mechanism for recovering part of it, confirming that the bottleneck is representational rather than architectural.

\section*{Acknowledgements}
The author thanks Peter Lukianchenko (Faculty of Computer Science, HSE
University) for supervising this research project. The problem studied here
originated in a production CTC-ASR setting encountered during an industry
internship.

\bibliography{references}

\begin{thebibliography}{33}
\providecommand{\natexlab}[1]{#1}
\providecommand{\url}[1]{\texttt{#1}}
\expandafter\ifx\csname urlstyle\endcsname\relax
  \providecommand{\doi}[1]{doi: #1}\else
  \providecommand{\doi}{doi: \begingroup \urlstyle{rm}\Url}\fi

\bibitem[Casella \& Robert(1996)Casella and Robert]{casella1996rao}
Casella, G. and Robert, C.~P.
\newblock {Rao--Blackwellisation} of sampling schemes.
\newblock \emph{Biometrika}, 83\penalty0 (1):\penalty0 81--94, 1996.

\bibitem[Eikema \& Aziz(2020)Eikema and Aziz]{eikema2020map}
Eikema, B. and Aziz, W.
\newblock Is {MAP} decoding all you need? the inadequacy of the mode in neural
  machine translation.
\newblock In \emph{Proceedings of the 28th International Conference on
  Computational Linguistics (COLING)}, pp.\  4506--4520, 2020.

\bibitem[Finkelstein et~al.(2024)Finkelstein, Naskar, Mirzazadeh, Shah, and
  Freitag]{finkelstein2024mbr}
Finkelstein, M., Naskar, S., Mirzazadeh, M., Shah, A., and Freitag, M.
\newblock {MBR} and {QE} finetuning: Training-time distillation of the best and
  most expensive decoding methods.
\newblock In \emph{International Conference on Learning Representations}, 2024.
\newblock arXiv:2309.10966.

\bibitem[Freitag et~al.(2022)Freitag, Grangier, Tan, and
  Liang]{freitag2022high}
Freitag, M., Grangier, D., Tan, Q., and Liang, B.
\newblock High-quality rather than high model probability: Minimum {Bayes} risk
  decoding with neural metrics.
\newblock \emph{Transactions of the Association for Computational Linguistics},
  10:\penalty0 811--825, 2022.

\bibitem[Goel \& Byrne(2000)Goel and Byrne]{goel2000mbr}
Goel, V. and Byrne, W.~J.
\newblock Minimum {Bayes}-risk automatic speech recognition.
\newblock \emph{Computer Speech \& Language}, 14\penalty0 (2):\penalty0
  115--135, 2000.

\bibitem[Graves \& Jaitly(2014)Graves and Jaitly]{graves2014towards}
Graves, A. and Jaitly, N.
\newblock Towards end-to-end speech recognition with recurrent neural networks.
\newblock In \emph{Proceedings of the 31st International Conference on Machine
  Learning (ICML)}, pp.\  1764--1772, 2014.

\bibitem[Graves et~al.(2006)Graves, Fern{\'a}ndez, Gomez, and
  Schmidhuber]{graves2006ctc}
Graves, A., Fern{\'a}ndez, S., Gomez, F., and Schmidhuber, J.
\newblock Connectionist temporal classification: Labelling unsegmented sequence
  data with recurrent neural networks.
\newblock In \emph{Proceedings of the 23rd International Conference on Machine
  Learning (ICML)}, pp.\  369--376, 2006.

\bibitem[G{\"u}l{\c{c}}ehre et~al.(2015)G{\"u}l{\c{c}}ehre, Firat, Xu, Cho,
  Barrault, Lin, Bougares, Schwenk, and Bengio]{gulcehre2015fusion}
G{\"u}l{\c{c}}ehre, {\c{C}}., Firat, O., Xu, K., Cho, K., Barrault, L., Lin,
  H.-C., Bougares, F., Schwenk, H., and Bengio, Y.
\newblock On using monolingual corpora in neural machine translation.
\newblock \emph{arXiv preprint arXiv:1503.03535}, 2015.

\bibitem[Hernandez et~al.(2018)Hernandez, Nguyen, Ghannay, Tomashenko, and
  Est{\`e}ve]{hernandez2018tedlium3}
Hernandez, F., Nguyen, V., Ghannay, S., Tomashenko, N., and Est{\`e}ve, Y.
\newblock {TED-LIUM 3}: Twice as much data and corpus repartition for
  experiments on speaker adaptation.
\newblock \emph{arXiv preprint arXiv:1805.04699}, 2018.

\bibitem[Hori et~al.(2017)Hori, Watanabe, and Hershey]{hori2017advances}
Hori, T., Watanabe, S., and Hershey, J.~R.
\newblock Joint {CTC/Attention} decoding for end-to-end speech recognition.
\newblock In \emph{Proceedings of the 55th Annual Meeting of the Association
  for Computational Linguistics (ACL)}, pp.\  518--529, 2017.

\bibitem[Huang et~al.(2024)Huang, Ye, Tan, and Li]{huang2024crctc}
Huang, R., Ye, Z., Tan, T., and Li, J.
\newblock {CR-CTC}: Consistency regularization on {CTC} for improved speech
  recognition.
\newblock \emph{arXiv preprint arXiv:2407.21188}, 2024.

\bibitem[Kuang(2022)]{kaldifeat}
Kuang, F.
\newblock kaldifeat: A {Python} library for computing {Kaldi}-compatible
  acoustic features, 2022.
\newblock Open-source library. \url{https://github.com/csukuangfj/kaldifeat}.

\bibitem[Kuang et~al.(2022)Kuang, Povey, Wu, Kang, Yao, and Yang]{li2022k2}
Kuang, F., Povey, D., Wu, L., Kang, W., Yao, Z., and Yang, X.
\newblock k2: {GPU}-accelerated {CTC} lattice beam search and {N}-best
  extraction, 2022.
\newblock Open-source library. \url{https://github.com/k2-fsa/k2}.

\bibitem[Kumar \& Byrne(2004)Kumar and Byrne]{kumar2004mbr}
Kumar, S. and Byrne, W.
\newblock Minimum {Bayes}-risk decoding for statistical machine translation.
\newblock In \emph{Human Language Technology Conference and North American
  Chapter of the Association for Computational Linguistics (HLT-NAACL)}, pp.\
  169--176, 2004.

\bibitem[Liu et~al.(2019)Liu, Ott, Goyal, Du, Joshi, Chen, Levy, Lewis,
  Zettlemoyer, and Stoyanov]{liu2019roberta}
Liu, Y., Ott, M., Goyal, N., Du, J., Joshi, M., Chen, D., Levy, O., Lewis, M.,
  Zettlemoyer, L., and Stoyanov, V.
\newblock {RoBERTa}: A robustly optimized {BERT} pretraining approach.
\newblock \emph{arXiv preprint arXiv:1907.11692}, 2019.

\bibitem[Miao et~al.(2015)Miao, Gowayyed, and Metze]{miao2015eesen}
Miao, Y., Gowayyed, M., and Metze, F.
\newblock {EESEN}: End-to-end speech recognition using deep {RNN} models and
  {WFST}-based decoding.
\newblock In \emph{2015 IEEE Workshop on Automatic Speech Recognition and
  Understanding (ASRU)}, pp.\  167--174, 2015.

\bibitem[Mikolov et~al.(2010)Mikolov, Karafiat, Burget, {\v{C}}ernock{\'y}, and
  Khudanpur]{mikolov2010recurrent}
Mikolov, T., Karafiat, M., Burget, L., {\v{C}}ernock{\'y}, J., and Khudanpur,
  S.
\newblock Recurrent neural network based language model.
\newblock In \emph{Proceedings of Interspeech 2010}, pp.\  1045--1048, 2010.

\bibitem[Panayotov et~al.(2015)Panayotov, Chen, Povey, and
  Khudanpur]{panayotov2015librispeech}
Panayotov, V., Chen, G., Povey, D., and Khudanpur, S.
\newblock {LibriSpeech}: An {ASR} corpus based on public domain audio books.
\newblock In \emph{2015 IEEE International Conference on Acoustics, Speech and
  Signal Processing (ICASSP)}, pp.\  5206--5210, 2015.

\bibitem[Park et~al.(2019)Park, Chan, Zhang, Chiu, Zoph, Cubuk, and
  Le]{park2019specaugment}
Park, D.~S., Chan, W., Zhang, Y., Chiu, C.-C., Zoph, B., Cubuk, E.~D., and Le,
  Q.~V.
\newblock {SpecAugment}: A simple data augmentation method for automatic speech
  recognition.
\newblock In \emph{Proceedings of Interspeech 2019}, pp.\  2613--2617, 2019.

\bibitem[Prabhavalkar et~al.(2018)Prabhavalkar, Sainath, Wu, Nguyen, Chen,
  Kannan, and Narayanan]{prabhavalkar2018mwer}
Prabhavalkar, R., Sainath, T.~N., Wu, Y., Nguyen, P., Chen, Z., Kannan, A., and
  Narayanan, A.
\newblock Minimum word error rate training for attention-based
  sequence-to-sequence models.
\newblock In \emph{2018 IEEE International Conference on Acoustics, Speech and
  Signal Processing (ICASSP)}, pp.\  4839--4843, 2018.

\bibitem[Radford et~al.(2019)Radford, Wu, Child, Luan, Amodei, and
  Sutskever]{radford2019gpt2}
Radford, A., Wu, J., Child, R., Luan, D., Amodei, D., and Sutskever, I.
\newblock Language models are unsupervised multitask learners.
\newblock \emph{OpenAI Blog}, 1\penalty0 (8):\penalty0 9, 2019.

\bibitem[Ranganath et~al.(2014)Ranganath, Gerrish, and Blei]{ranganath2014bbvi}
Ranganath, R., Gerrish, S., and Blei, D.~M.
\newblock Black box variational inference.
\newblock In \emph{Proceedings of the 17th International Conference on
  Artificial Intelligence and Statistics (AISTATS)}, pp.\  814--822, 2014.

\bibitem[Rennie et~al.(2017)Rennie, Marcheret, Mroueh, Ross, and
  Goel]{rennie2017self}
Rennie, S.~J., Marcheret, E., Mroueh, Y., Ross, J., and Goel, V.
\newblock Self-critical sequence training for image captioning.
\newblock In \emph{2017 IEEE Conference on Computer Vision and Pattern
  Recognition (CVPR)}, pp.\  7008--7024, 2017.

\bibitem[Salazar et~al.(2020)Salazar, Liang, Nguyen, and
  Kirchhoff]{salazar2020pll}
Salazar, J., Liang, D., Nguyen, T.~Q., and Kirchhoff, K.
\newblock Masked language model scoring.
\newblock In \emph{Proceedings of the 58th Annual Meeting of the Association
  for Computational Linguistics (ACL)}, pp.\  2699--2712, 2020.

\bibitem[Sanh et~al.(2019)Sanh, Debut, Chaumond, and Wolf]{sanh2019distilbert}
Sanh, V., Debut, L., Chaumond, J., and Wolf, T.
\newblock {DistilBERT}, a distilled version of {BERT}: Smaller, faster, cheaper
  and lighter.
\newblock \emph{arXiv preprint arXiv:1910.01108}, 2019.

\bibitem[Shannon(2017)]{shannon2017}
Shannon, M.
\newblock Optimizing expected word error rate via sampling for speech
  recognition.
\newblock In \emph{Proceedings of Interspeech 2017}, pp.\  3953--3957, 2017.

\bibitem[Shao et~al.(2024)Shao, Wang, Zhu, Xu, Song, Bi, Zhang, Zhang, Li, Wu,
  and Guo]{shao2024grpo}
Shao, Z., Wang, P., Zhu, Q., Xu, R., Song, J., Bi, X., Zhang, H., Zhang, M.,
  Li, Y.~K., Wu, Y., and Guo, D.
\newblock {DeepSeekMath}: Pushing the limits of mathematical reasoning in open
  language models.
\newblock \emph{arXiv preprint arXiv:2402.03300}, 2024.

\bibitem[Shin et~al.(2019)Shin, Lee, and Jung]{shin2019effective}
Shin, J.~S., Lee, Y., and Jung, K.
\newblock Effective sentence scoring method using {BERT} for speech
  recognition.
\newblock In \emph{Proceedings of the Asian Conference on Machine Learning
  (ACML)}, pp.\  1081--1093, 2019.

\bibitem[Skalse et~al.(2022)Skalse, Howe, Krasheninnikov, and
  Krueger]{skalse2022rewardhacking}
Skalse, J., Howe, N., Krasheninnikov, D., and Krueger, D.
\newblock Defining and characterizing reward hacking.
\newblock In \emph{Advances in Neural Information Processing Systems
  (NeurIPS)}, volume~35, pp.\  9460--9471, 2022.

\bibitem[Snyder et~al.(2015)Snyder, Chen, and Povey]{snyder2015musan}
Snyder, D., Chen, G., and Povey, D.
\newblock {MUSAN}: A music, speech, and noise corpus, 2015.
\newblock arXiv:1510.08484.

\bibitem[Wang et~al.(2021)Wang, Rivi{\`e}re, Lee, Wu, Talnikar, Haziza,
  Williamson, Pino, and Dupoux]{wang2021voxpopuli}
Wang, C., Rivi{\`e}re, M., Lee, A., Wu, A., Talnikar, C., Haziza, D.,
  Williamson, M., Pino, J., and Dupoux, E.
\newblock {VoxPopuli}: A large-scale multilingual speech corpus for
  representation learning, semi-supervised learning and interpretation.
\newblock In \emph{Proceedings of the 59th Annual Meeting of the Association
  for Computational Linguistics (ACL)}, pp.\  993--1003, 2021.

\bibitem[Williams(1992)]{williams1992}
Williams, R.~J.
\newblock Simple statistical gradient-following algorithms for connectionist
  reinforcement learning.
\newblock \emph{Machine Learning}, 8\penalty0 (3--4):\penalty0 229--256, 1992.

\bibitem[Yao et~al.(2024)Yao, Wu, Kuang, Kang, Yang, Yang, Lin, and
  Povey]{yao2024zipformer}
Yao, Z., Wu, L., Kuang, F., Kang, W., Yang, X., Yang, Y., Lin, L., and Povey,
  D.
\newblock Zipformer: A faster and better encoder for automatic speech
  recognition.
\newblock In \emph{The Twelfth International Conference on Learning
  Representations (ICLR)}, 2024.

\end{thebibliography}
\bibliographystyle{icml2025}

\appendix
\section{Full Results Tables}

This appendix presents extended versions of the two inline tables from the main text and the
complete MWER training trajectories for all five model configurations.

\subsection{Master Results Table}

Table~\ref{tab:master-full} extends Table~\ref{tab:cross-condition} in §6.1 with 95\% confidence
intervals. All results use MBR-CER + RoBERTa PLL ($\tau$=10) with zero retuning across conditions.
Statistics: paired bootstrap, $B$=10\,000, seed=42, all comparisons against greedy decoding. Gap
closure is $(W_{\text{greedy}} - W_{\text{MBR}}) / (W_{\text{greedy}} - W_{\text{oracle}})$.
VoxPopuli WERs are punctuation-stripped (§B.3); raw values are greedy 21.93\%, oracle 21.51\%,
MBR 21.97\%.

\begin{table*}[t]
\centering
\footnotesize
\caption{Full results: MBR-CER\,+\,RoBERTa PLL ($\tau$=10) vs greedy across all evaluation
conditions, with 95\% bootstrap confidence intervals. Paired bootstrap, $B$=10\,000, seed=42.
n.s.\ = $p>0.05$. VoxPopuli punct-stripped (§B.3).
The test-other $G$=16 condition (§6.1) is reported inline only and not duplicated here.
$^{\dagger}$MUSAN 0\,dB: the corpus-level WER delta is driven by a few high-error utterances; the
paired bootstrap CI reflects the per-utterance resampling distribution and shows no consistent direction.}
\label{tab:master-full}
\begin{tabular}{llrrrrrrr}
\hline
Condition & $G$ & Greedy & Oracle & MBR+PLL & $\Delta$ & $p$ & 95\% CI & Gap cl. \\
 & & (\%) & (\%) & (\%) & (pp) & & (pp) & (\%) \\
\hline
\multicolumn{9}{l}{\textit{In-domain — LibriSpeech}} \\
dev-clean    & 16  & 2.37  & 1.54  & 2.28  & $-$0.09 & 0.008    & [$-$0.16,\;$-$0.02] & 10.3 \\
dev-other    & 16  & 6.02  & 4.44  & 5.79  & $-$0.23 & $<$0.0001 & [$-$0.33,\;$-$0.14] & 14.7 \\
dev-other    & 128 & 6.02  & 3.53  & 5.53  & $-$0.49 & $<$0.0001 & [$-$0.59,\;$-$0.40] & 19.8 \\
test-other   & 128 & 5.96  & 3.37  & 5.42  & $-$0.54 & $<$0.0001 & [$-$0.63,\;$-$0.44] & 20.7 \\
\hline
\multicolumn{9}{l}{\textit{Cross-architecture — Zipformer-M 65M (LibriSpeech dev-other)}} \\
Zipformer-M  & 16  & 4.78  & 3.44  & 4.56  & $-$0.22 & $<$0.0001 & [$-$0.30,\;$-$0.14] & 16.5 \\
Zipformer-M  & 128 & 4.78  & 2.73  & 4.43  & $-$0.34 & $<$0.0001 & [$-$0.42,\;$-$0.27] & 16.8 \\
\hline
\multicolumn{9}{l}{\textit{Out-of-domain}} \\
TED-LIUM 3   & 16  & 11.30 & 9.16  & 10.97 & $-$0.33 & $<$0.0001 & [$-$0.50,\;$-$0.17] & 15.6 \\
TED-LIUM 3   & 128 & 11.30 & 7.51  & 10.57 & $-$0.73 & $<$0.0001 & [$-$0.88,\;$-$0.58] & 19.2 \\
VoxPopuli    & 128 & 18.29 & 17.97 & 18.33 & $+$0.04 & n.s.       & ---                  & ---  \\
\hline
\multicolumn{9}{l}{\textit{Noise robustness — MUSAN additive noise on LibriSpeech dev-other ($G$=16)}} \\
MUSAN 20\,dB & 16  & 6.62  & 4.89  & 6.38  & $-$0.24 & 0.006 & [$-$0.25,\;$-$0.03] & 13.9 \\
MUSAN 10\,dB & 16  & 8.04  & 6.26  & 7.80  & $-$0.24 & 0.003 & [$-$0.28,\;$-$0.05] & 13.5 \\
MUSAN\;\,5\,dB  & 16  & 11.10 & 9.06  & 10.84 & $-$0.27 & 0.001 & [$-$0.29,\;$-$0.07] & 13.0 \\
MUSAN\;\,0\,dB$^{\dagger}$  & 16  & 17.88 & 15.57 & 17.64 & $-$0.23 & 0.646 & [$-$0.10,\;$+$0.14] & ---  \\
\hline
\end{tabular}
\end{table*}

Eleven of 13 conditions are significant at $\alpha$=0.05. VoxPopuli fails through coverage:
91.5\% of utterances are already greedy-optimal in the $G$=128 candidate set (oracle gap 0.31~pp),
so no amount of reranking can help. At MUSAN~0~dB, candidate quality degrades under extreme acoustic noise; p=0.646 despite
$\Delta$=$-$0.24~pp indicates that the per-utterance gain variance is very high. Both failures were
predicted in §5.3 by the Spearman~$\rho$ divergence analysis.

All confidence intervals: paired bootstrap, $B=10{,}000$, seed=42.

\subsection{CTC-Internal Methods: Extended Table with Confidence Intervals}

Table~\ref{tab:ctc-internal-full} reproduces the §5.1 inline table (Table~\ref{tab:ctc-internal})
and adds 95\% bootstrap confidence intervals where available. Bootstrap CIs use
$B$=10\,000, seed=42, $G$=16, $n$=2864 utterances. Eight methods were included in the primary bootstrap run; the remaining
five are from individual experiment reports and do not have CIs.

\begin{table*}[t]
\centering
\small
\caption{CTC-internal and acoustic-feature scoring methods at $G$=16, LibriSpeech dev-other
(greedy WER 6.022\%). Paired bootstrap $B$=10\,000, seed=42. CI available for methods in the
primary bootstrap run only. ${}^{*}$MBR-WER uses WER as both training utility and evaluation
metric — result is circular; see §6.2 for discussion.}
\label{tab:ctc-internal-full}
\begin{tabular}{lrrrr}
\hline
Method & WER (\%) & $\Delta$ (pp) & $p$ & 95\% CI (pp) \\
\hline
Greedy (baseline)                     & 6.022         & 0.000    & ---       & --- \\
Argmax $P_{\text{CTC}}$               & 6.022         & 0.000    & 1.000     & [0.00,\;0.00] \\
Length-norm (tokens)                   & 6.022         & 0.000    & 0.652     & [$-$0.01,\;$+$0.01] \\
Length-norm (chars)                    & 6.022         & 0.000    & 1.000     & [0.00,\;0.00] \\
MBR-WER ($\tau$=1)$^{*}$             & 6.022         & 0.000    & 1.000     & [0.00,\;0.00] \\
MBR-CER ($\tau$=1)                    & 6.028         & $+$0.006 & 1.000     & [0.00,\;$+$0.02] \\
Self-consistency (uniform)             & 6.040         & $+$0.018 & 0.721     & [$-$0.05,\;$+$0.08] \\
MBR-CER ($\tau$=50, best CTC-only)    & 5.987         & $-$0.035 & 0.163     & [$-$0.10,\;$+$0.03] \\
\hline
Contrastive decoding ($\alpha$=0.1)   & 6.071         & $+$0.049 & ---       & --- \\
MC-dropout ($T$=4, 5 seeds)           & $6.030\pm0.020$ & $+$0.008 & 0/5 sig.\ & --- \\
14-feature MLP rescorer               & 6.050         & $+$0.028 & ---       & --- \\
3-gram shallow fusion ($\alpha$=0.9)  & 6.018         & $-$0.004 & 0.368     & --- \\
Encoder value head                    & 6.020         & $-$0.002 & ---       & --- \\
\hline
\end{tabular}
\end{table*}

The confidence intervals make the null result precise: every method with a CI covering zero
cannot be distinguished from greedy at any standard significance threshold. MBR-CER at $\tau$=50
is the strongest CTC-only result ($\Delta$=$-$0.035~pp, 95\%\ CI [$-$0.10, $+$0.03]) yet
remains non-significant. The encoder value head and 3-gram results lack CIs but have $p$$>$0.1
from their individual reports. The contrastive, MLP, and MC-dropout methods were not included in
the primary bootstrap because corpus-level metrics (contrastive, MLP) or seed-level variance
(MC-dropout) make a single bootstrap uninformative.

\subsection{MWER Training Trajectories}

Tables~\ref{tab:mwer-crctc} and~\ref{tab:mwer-stdctc} give complete training trajectories for
all MWER and distillation runs. All CR-CTC experiments start from the same frozen checkpoint
(training-pipeline baseline 6.67\% on dev-other; see §4.1 for the 6.67\%\,vs.\,6.02\%
discrepancy due to fbank computation). The standard CTC experiment starts from a different
checkpoint (training-pipeline baseline 7.07\%). All $\Delta$ values are computed within-pipeline
against each experiment's own baseline.

\begin{table*}[t]
\centering
\small
\caption{CR-CTC MWER training trajectories (dev-other WER). All four configurations start from
the same baseline (6.67\%). Subset configurations (MWER-unclipped-subset, MWER-clipped-subset): 10 epochs on train-clean-100 subset (${\sim}248$ steps/epoch). Full-data configurations (MWER-unclipped-full, MWER-clipped-full): 1 epoch on full train-clean-100 (7132 steps). Clipped = GRPO-style importance-ratio clipping. Baseline $\Delta$=0 by definition.}
\label{tab:mwer-crctc}
\begin{tabular}{lrrrr}
\hline
Epoch & Unc.-subset & Clip.-subset & Unc.-full & Clip.-full \\
 & WER\,(\%) & WER\,(\%) & WER\,(\%) & WER\,(\%) \\
\hline
0 (baseline) & 6.67 & 6.67 & 6.67 & 6.67 \\
1            & 8.53 & 8.48 & 15.30 & 15.57 \\
2            & 8.31 & 8.94 & ---   & ---   \\
3            & 9.22 & 9.31 & ---   & ---   \\
4            & 9.62 & 9.80 & ---   & ---   \\
5            & 9.99 & 10.44 & ---   & ---   \\
6            & 11.11 & 10.95 & ---   & ---   \\
7            & 11.29 & 11.55 & ---   & ---   \\
8            & 12.32 & 12.49 & ---   & ---   \\
9            & 13.43 & 13.31 & ---   & ---   \\
10           & 13.49 & 12.85 & ---   & ---   \\
\hline
$\Delta$ final & $+$6.82\,pp & $+$6.18\,pp & $+$8.63\,pp & $+$8.90\,pp \\
\hline
\end{tabular}
\end{table*}

\begin{table}[h!]
\centering
\small
\caption{Standard CTC MWER trajectory (dev-other WER, ${\sim}$40\% of one epoch). Baseline
7.07\%; all evaluations use the same greedy decoding pipeline. The monotonic drift is qualitatively
distinct from the CR-CTC catastrophic collapse: no phase transition, linear increase, endpoint
+3.4\% relative.}
\label{tab:mwer-stdctc}
\begin{tabular}{rrr}
\hline
Step & WER (\%) & $\Delta$ (pp) \\
\hline
0    & 7.07 & $+$0.00 \\
500  & 7.12 & $+$0.05 \\
1000 & 7.14 & $+$0.07 \\
1500 & 7.21 & $+$0.14 \\
2000 & 7.25 & $+$0.18 \\
2500 & 7.31 & $+$0.24 \\
3000 & 7.31 & $+$0.24 \\
\hline
\end{tabular}
\end{table}

The CR-CTC trajectories (Table~\ref{tab:mwer-crctc}) share two properties: monotonic increase from
epoch~1 and no sign of stabilization through epoch~10. Clipping (MWER-clipped-subset vs MWER-unclipped-subset) reduces the
epoch-10 WER by 0.64~pp (12.85 vs 13.49\%) but does not flatten the trajectory or produce any
epoch with lower WER than the previous one. Full-data runs (MWER-unclipped-full and MWER-clipped-full) collapse more severely
in one epoch than the subset runs do in ten, confirming that more gradient steps accelerate the
failure rather than correcting it.

The standard CTC trajectory (Table~\ref{tab:mwer-stdctc}) drifts linearly at approximately
0.08~pp per 1000 steps, reaching $+$0.24~pp total over 3000 steps. This behavior — slow linear
drift without catastrophic phase transition — distinguishes the standard CTC failure mode (basin
geometry, §4.4) from the CR-CTC failure mode (missing reward signal, §4.2).

\section{Supplementary Experiments}

This appendix documents three methodological decisions that affect numerical results in the main
text: the canonical N-best generation configuration, the sensitivity of oracle quality to the
\texttt{nbest\_scale} sampling parameter, and the punctuation correction applied to VoxPopuli.

\subsection{Canonical N-best Generation Configuration}

All N-best lists in this work are generated by a single pipeline: k2 lattice sampling
\citep{li2022k2} from the frozen CTC model at \texttt{nbest\_scale}=1.0 with $64\times$
oversampling, followed by deduplication to the target beam size $G$. The N-best file for the
primary dev-other experiments is \texttt{nbest\_dev\_other\_G16.jsonl}. Under this configuration
the oracle WER on LibriSpeech dev-other is \textbf{4.44\%} at $G$=16 and \textbf{3.53\%} at
$G$=128; these are the ceilings against which every gap-closure percentage in the main text is
computed. The beam-size scaling curve (§5.3, Fig.~\ref{fig:gscaling}) draws a fresh N-best list
at each $G \in \{4, 8, 16, 32, 64, 128\}$ from the same lattices under the same settings.

N-best generation is effectively deterministic. Three seeds (42, 137, 2024) applied via
\texttt{torch.manual\_seed} before each \texttt{k2.Nbest.from\_lattice} call produced
byte-identical N-best lists at $G$=16: the cross-seed standard deviation of both oracle and MBR
WER was exactly 0.0000\,pp. The k2 \texttt{random\_paths} function uses an internal RNG decoupled
from PyTorch, and at \texttt{nbest\_scale}=1.0 the posterior is peaked enough that the sampled
path set saturates for most utterances given a fixed lattice and k2 version. Bootstrap confidence
intervals are therefore the only relevant source of uncertainty for the decode-time results.

\subsection{\texttt{nbest\_scale} Sensitivity}

\texttt{k2.Nbest.from\_lattice} accepts an \texttt{nbest\_scale} parameter $s > 0$ that scales
all lattice arc scores by $1/s$ before sampling. At $s$=1.0, paths are sampled in proportion
to their original lattice probabilities, so candidates concentrate near the most probable paths.
At $s < 1.0$, the distribution flattens: paths scatter more uniformly across the lattice.

The practical effect is large.

\begin{table*}[t]
\centering
\small
\caption{Effect of \texttt{nbest\_scale} on G=16 N-best oracle quality (LibriSpeech dev-other,
$n$=2864). Same model, same data, same oversample=64; only \texttt{nbest\_scale} varies.
Oracle gap = $(W_{\text{greedy}} - W_{\text{oracle}}) / W_{\text{greedy}}$.}
\label{tab:nbest-scale}
\begin{tabular}{rrrrrr}
\hline
\texttt{nbest\_scale} & Oracle WER (\%) & Oracle gap (\%) & Recoverable utts & Unique cands/utt & Mean pw. edit dist. \\
\hline
1.00 & 4.44 & 26.2 & 665 (23.2\%) & 15.5 & 2.65 \\
0.75 & 5.18 & 13.9 & 375 (13.1\%) & 16.0 & 3.82 \\
0.50 & 5.86 & \phantom{0}2.6 & \phantom{0}74\phantom{0} (2.6\%) & 16.0 & 8.24 \\
\hline
\end{tabular}
\end{table*}

Reducing \texttt{nbest\_scale} from 1.0 to 0.5 destroys 90\% of the oracle gap (26.2\% $\to$
2.6\%) while leaving candidate count almost unchanged (15.5 $\to$ 16.0 unique hypotheses per
utterance). The high mean pairwise edit distance at \texttt{scale}=0.5 (8.24 vs 2.65) confirms
that the candidates differ at many token positions, but those differences are at positions where
greedy was already correct. The candidates are lexically diverse but not functionally so: sixteen
unique strings, all roughly as good as — or worse than — greedy.

\textbf{Why this matters.}  The TED-LIUM~3 cross-domain and MUSAN noise robustness experiments originally used \texttt{nbest\_scale}=0.5, copied from an early exploratory file.
With that setting, the oracle gaps were 1.2\% (TL3) and 0.9\% (MUSAN) — suppressed by an order
of magnitude. Both experiments were rerun with \texttt{nbest\_scale}=1.0; the TL3 result in
§6.1 (oracle WER 9.16\%, absolute gap 2.14~pp at $G$=16) and the MUSAN results (oracle gap
13.0–13.9\% across 5–20~dB) reflect the corrected generation.

\textbf{Why \texttt{scale}=0.5 generates more edit distance.}  At $s$=0.5, the effective
temperature of the sampling distribution doubles. Paths that differ from the best path by a
large score penalty (misrecognized words) become almost as likely to be sampled as paths close
to the best path. These high-penalty paths produce strings that differ from greedy at the
misrecognized positions — exactly where the model is most confident about its (correct) output.
The resulting diversity is adversarial rather than useful.

\textbf{Canonical value.}  \texttt{nbest\_scale}=1.0 is the only setting used in any result
reported in the main text. All scripts assert this value before generating N-best lists.

\subsection{VoxPopuli Punctuation Correction}

The BPE-500 tokenizer used for all Zipformer-S and Zipformer-M experiments has no punctuation
tokens in its vocabulary. The model cannot produce commas, full stops, hyphens, or question marks.
VoxPopuli reference transcripts, which are drawn from European Parliament proceedings, contain
substantial punctuation.

Computing WER against punctuated references inflates all three metrics (greedy, oracle, MBR)
uniformly, since the model can never match the punctuation tokens that appear in references.
The effect is approximately 3.6~pp on this corpus.

\begin{table*}[t]
\centering
\small
\caption{VoxPopuli ($n$=1842 utterances, $G$=128) before and after punctuation stripping.
Regex: remove \texttt{[.,;?!:"$-$]} then collapse whitespace; apostrophes preserved.
$\Delta_{\text{oracle}}$ = greedy $-$ oracle; $\Delta_{\text{MBR}}$ = MBR $-$ greedy.}
\label{tab:voxpopuli-punct}
\begin{tabular}{lrrrr}
\hline
Metric & Raw WER (\%) & Punct-stripped WER (\%) & Correction (pp) \\
\hline
Greedy     & 21.93 & 18.29 & $-$3.64 \\
Oracle     & 21.51 & 17.97 & $-$3.54 \\
MBR+PLL    & 21.97 & 18.33 & $-$3.64 \\
\hline
Oracle gap ($\Delta_\text{oracle}$)  & 0.42 & 0.31 & --- \\
MBR vs greedy ($\Delta_\text{MBR}$) & $+$0.04 & $+$0.04 & --- \\
\hline
\end{tabular}
\end{table*}

The correction is uniform across all three metrics ($\pm$0.10~pp): the BPE-500 model produces no
punctuation in any hypothesis, so every hypothesis is missing exactly the same punctuation tokens
that appear in every reference. The oracle gap and MBR delta are therefore unaffected
($+$0.04~pp before and after stripping).

At 3.6~pp, the magnitude is large enough to matter. It accounts for 16.4\% of the raw greedy
WER and would make VoxPopuli appear harder than it is relative to LibriSpeech (where BPE-500
transcripts happen to contain no punctuation mismatch) if raw values were used.

The coverage-collapse conclusion is unaffected by the correction. The corrected oracle gap is
0.31~pp (18.29 $-$ 17.97), versus 2.49~pp on LibriSpeech dev-other at $G$=128 — an order of
magnitude smaller. 91.5\% of VoxPopuli utterances are already greedy-optimal in the $G$=128
candidate set; the failure of MBR on VoxPopuli is a coverage problem, not a scoring problem.

All VoxPopuli WERs in the main text (§6.1 and Table~\ref{tab:master-full}) use the
punctuation-stripped values: greedy 18.29\%, oracle 17.97\%, MBR 18.33\%.

\section{Detailed Method Discussions}

This appendix contains extended discussions of methods summarized in the main text. All results use the LibriSpeech dev-other evaluation set ($n$=2864 utterances) unless otherwise noted.

\subsection{CTC-Internal Scoring Method Details}

The following three methods from §\ref{sec:exhaustion} (Table~\ref{tab:ctc-internal}) produced null results. Extended analysis is provided here; the main text reports one-sentence summaries.

\subsubsection{MC-dropout analysis}

An initial exploratory run found a single seed with MC-MBR
WER 5.98\% — the best figure among all CTC-internal methods, apparently closing 10.3\% of the oracle
gap. Five independent seeds were then run (seed $\in$ \{42, 123, 456, 789,
1024\}) with $B$=10\,000 bootstrap for each. Zero of five seeds reached significance at $\alpha$=0.05;
$p$-values ranged from 0.27 to 0.77; three of five seeds produced WER above greedy. The
Zipformer-S architecture contains one dropout layer in the CTC projection head ($p$=0.1). All encoder
computation is deterministic at inference. Averaging $T$ stochastic projections of the same
deterministic encoder output yields marginal posterior smoothing, but no new information enters the
system. The effect direction is inconsistent across seeds; the null result holds.

\subsubsection{Contrastive decoding on CR-CTC}

The contrastive decoding variant compared the model's standard forward pass against a run
on SpecAugment-masked \cite{park2019specaugment} input features, subtracting the masked log-probabilities from the standard
ones. Every value $\alpha > 0$ degraded WER monotonically, reaching 7.14\% at $\alpha$=1.0. CR-CTC
is specifically trained for output consistency under augmentation \cite{huang2024crctc}: the masked
run therefore does not behave as a qualitatively different ``amateur'' that makes distinct errors. It
produces the same predictions with added noise. Subtracting noise from signal reduces the signal, and
the blank token — dominant in CTC frame distributions and most consistent under masking — is penalised
disproportionately. This is a fundamental incompatibility between the contrastive method's premise
and CR-CTC's training objective, not a tuning failure.

\subsubsection{MLP rescorer}

The 14-feature MLP rescorer used CTC log-probability variants, hypothesis length
statistics, and inter-candidate agreement scores. Its best configuration
produced WER 6.05\%. Cross-validation on train-clean-100 achieved $R^2$=0.76, confirming strong
in-distribution discriminability. The learned mapping did not transfer to per-utterance candidate
ranking on dev-other because the training criterion — predicting WER distribution across utterances —
does not match the evaluation criterion, which requires ranking within individual utterances.

\subsection{DistilBERT MWER Reranker Training Details}

A DistilBERT discriminative reranker \cite{sanh2019distilbert} was trained with MWER loss on LibriSpeech train-clean-100 (28,539 utterances $\times$ 16 candidates) and evaluated on dev-other at $G$=128.

\textbf{Training convergence.} MWER loss fell from 23.4~pp (random initialization) to 1.26~pp
over 16,050 steps, and held-out validation WER on train-clean-100 fell from 25.2\% to 1.22\%.
Training converged as expected on the training distribution.

\textbf{Evaluation.} On dev-other, every configuration using the trained reranker was statistically significantly
worse than the MBR-CER+PLL baseline ($p$=1.000 for all; 95\%\ CIs uniformly positive). The
optimal mixing weight in the MBR posterior sweep was $\beta$=0, meaning the system discarded the
reranker entirely and reverted to pure PLL. The best reranker-using configuration — three-way
argmax with CTC ($\alpha$=0.6), reranker ($\beta$=0.2), and PLL ($\gamma$=0.2) — produced 5.79\%,
trailing the MBR+PLL baseline by 0.26~pp.

\textbf{Mechanism.} On train-clean-100, greedy WER is 1.09\% and the greedy-oracle gap is 0.007~pp: there is
essentially no contrast between good and bad hypotheses for MWER to learn from. The reranker
learns to demote orthographically irregular candidates (the $G$=16 training N-best included
strings like ``afle shutot ran outed'') while recognizing acoustically realistic transcription
errors on dev-other as valid-looking text. PLL, trained on 160~GB of English, already captures
the available text-only discriminative signal. A DistilBERT model fine-tuned for five epochs on
28K ASR utterances without acoustic grounding adds nothing to that signal.

\end{document}